\documentclass[sigconf]{acmart}

\setcopyright{none}
\renewcommand\footnotetextcopyrightpermission[1]{}
\usepackage{soul}
\usepackage{hyperref}
\usepackage{multirow}
\usepackage[utf8]{inputenc}
\usepackage[ruled, lined, longend, linesnumbered]{algorithm2e}
\usepackage{amsmath,multicol}
\usepackage{subcaption,textcomp,comment}

\usepackage{booktabs}
\usepackage{multirow}

\usepackage{paralist}
\usepackage{pifont}

\usepackage{array}
\newcolumntype{L}[1]{>{\raggedright\let\newline\\\arraybackslash\hspace{0pt}}m{#1}}
\newcolumntype{C}[1]{>{\centering\let\newline\\\arraybackslash\hspace{0pt}}m{#1}}
\newcolumntype{R}[1]{>{\raggedleft\let\newline\\\arraybackslash\hspace{0pt}}m{#1}}

\newcommand{\mwx}[1]{\textbf{\color{red}{#1}}}
\newcommand{\sys}{\texttt{FedAdapter}\xspace}
\newcommand{\relativeacc}{relative target accuracy\xspace}

\usepackage{wrapfig}
\usepackage{comment}
\usepackage{colortbl}

\usepackage{hyperref}

\usepackage{color,soul}
\usepackage{graphics}
\usepackage{hyperref}
\usepackage{lipsum}
\usepackage{tikz}
\usepackage{enumitem}	
\usepackage{listings} 

\usepackage{booktabs}	
\usepackage{lipsum}

\usepackage{capt-of}	

\usepackage{todonotes}  
\usepackage{subcaption} 

\definecolor{refkey}{rgb}{0,0,1}
\definecolor{labelkey}{rgb}{0,0,1}

\usepackage{cleveref}
\AtBeginDocument{\DeclareCaptionSubType{lstlisting}}
\crefname{sublstlisting}{listing}{listings}
\Crefname{sublstlisting}{Listing}{Listings}

\usepackage{mathrsfs}	
\usepackage{amsfonts}

\usepackage[export]{adjustbox} 

\usepackage{boldline}					

\usepackage{multirow}

\renewcommand{\paragraph}[1]{\vskip 3pt\noindent\textbf{#1 }}	 

  {\begin{list}{$\bullet$}%
     {\setlength{\parsep}{0pt}%
      \setlength{\topsep}{0pt}%
      \setlength{\itemsep}{2pt}}}%
  {\end{list}}
\newcommand\Noted[1]{} 

\definecolor{darkblue}{rgb}{0.0, 0.0, 0.55}
\definecolor{mygreen}{HTML}{ADFF2F}
\definecolor{mylightgray}{gray}{0.8}

  {\begin{itemize}
	[leftmargin=0cm,
		itemindent=.3cm,
		labelwidth=\itemindent,
		labelsep=0pt,
		parsep=0.0pt,
		topsep=0.5pt,
		itemsep=0.5pt,
		align=left]
  }%
  {\end{itemize}}    

  {\begin{enumerate}
	[leftmargin=.cm,itemindent=.5cm,labelwidth=\itemindent,
		labelsep=0pt,
		parsep=1pt,
		topsep=1pt,
		itemsep=3pt,
		align=left]
  }%
  {\end{enumerate}}    

\makeatletter
\def\@copyrightspace{\relax}
\makeatother

\begin{document}

	\title[FedAdapter: Efficient Federated Learning for Modern NLP]
	{FedAdapter: Efficient Federated Learning for Modern NLP}

	\author{Dongqi Cai}
	\affiliation{
	\institution{Beiyou Shenzhen Institute}
	\country{}
	}

	\author{Yaozong Wu}
	\affiliation{
	\institution{Beiyou Shenzhen Institute}
	\country{}
	}

	\author{Shangguang Wang}
	\affiliation{
	\institution{Beiyou Shenzhen Institute}
	\country{}
	}

	\author{Felix Xiaozhu Lin}
	\affiliation{
	\institution{University of Virginia}
	\country{}
	}

	\author{Mengwei Xu}
	\affiliation{
	\institution{Beiyou Shenzhen Institute}
	\country{}
	}
	

	\begin{abstract}
Transformer-based pre-trained models have revolutionized NLP for superior performance and generality.
Fine-tuning pre-trained models for downstream tasks often requires private data, for which federated learning is the de-facto approach (i.e., FedNLP).
However, our measurements show that FedNLP is prohibitively slow due to the large model sizes and the resultant high network/computation cost.
Towards practical FedNLP, we identify as the key building blocks  \textit{adapters}, small bottleneck modules inserted at a variety of model layers.
A key challenge is to properly configure the depth and width of adapters, to which the training speed and efficiency is highly sensitive.
No silver-bullet configuration exists:  
the optimal choice varies across downstream NLP tasks,
desired model accuracy, 
and mobile resources.
To automate adapter configuration, 
we propose \sys\footnote[1]{
    \sys is available at \textit{\url{https://github.com/UbiquitousLearning/AdaFL}}.
    In the conference version (MobiCom'23) of this work, the system is named to \texttt{AdaFL} instead of \texttt{FedAdapter}.
}, a framework that enhances the existing FedNLP with two novel designs.
First, \sys progressively upgrades the adapter configuration throughout a training session; 
the principle is to quickly learn shallow knowledge by only training fewer and smaller adapters at the model's top layers, and incrementally learn deep knowledge by incorporating deeper and larger adapters.
Second, \sys continuously profiles future adapter configurations by allocating participant devices to trial groups.
Extensive experiments show that \sys can reduce FedNLP's model convergence delay to no more than several hours, which is up to 155.5$\times$ faster compared to vanilla FedNLP and 48$\times$ faster compared to strong baselines.

\end{abstract}
	
	\maketitle
	 


	\section{Introduction}\label{sec:intro}

\begin{figure}[t]
	\centering
	 \includegraphics[width=0.475\textwidth]{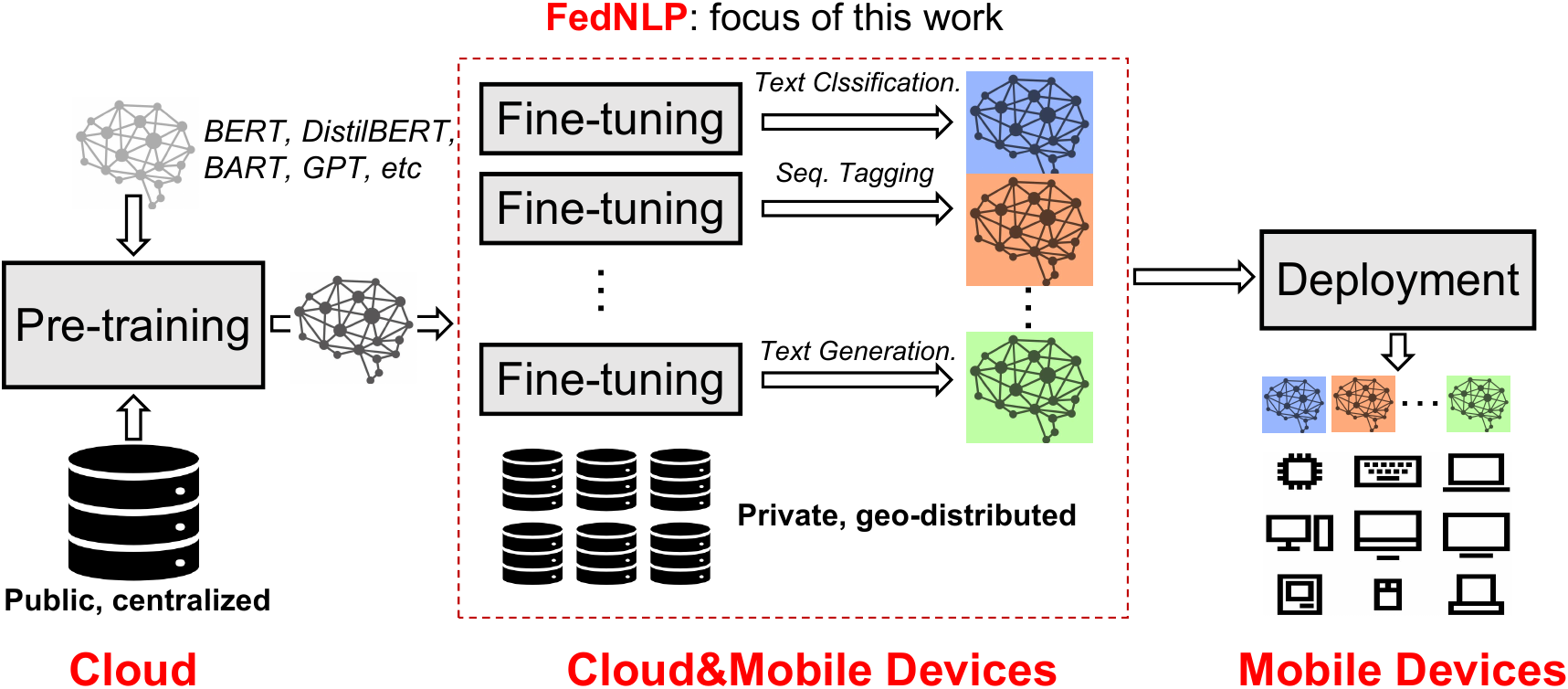}
	\caption{FedNLP and its role in modern NLP} 
	\label{fig:overview}
\end{figure}

With the recent rise of transformers and its  variants~\cite{vaswani2017attention,devlin2018bert,sanh2019distilbert, hou2020dynabert, liu2020fastbert, sun2020mobilebert, zafrir2019q8bert, bai2020binarybert}, 
modern NLP models show compelling use cases on mobile devices.
Examples include sentiment analysis, QA, and auto completion~\cite{zhang2018deep,shao2019transformer,van2019does,kim2021code, svyatkovskiy2020intellicode}.
Through careful engineering~\cite{sanh2019distilbert, hou2020dynabert, liu2020fastbert, sun2020mobilebert, bai2020binarybert}, inference with the NLP models is demonstrated to be affordable on mobile devices. 

Much success of modern NLP comes from its training workflow as illustrated in Figure~\ref{fig:overview}. 
(1) The pre-training phase initializes a model 
on large text corpora. 
The training is self-supervised and time-consuming, often taking hundreds if not thousands of GPU days~\cite{devlin2018bert,brown2020language}. 
Pre-training teaches the model a language's inherent structure, e.g. word distribution. 
(2) The fine-tuning phase further adapts a pre-trained model for a specific NLP task targeting a specific domain, e.g. to classify sentiments (a task) of user emails (a domain)~\cite{devlin2018bert}. 
Fine-tuning is indispensable to modern NLP training; 
only through it, the model maps the generic language understanding to the outputs for rich NLP tasks.

\paragraph{FedNLP}
The two NLP training phases require data of disparate natures. 
While pre-training is typically done on public text corpora such as Wikipedia articles, 
fine-tuning requires domain-specific samples such as user reviews, messages, or emails. 
In mobile computing, these samples are generated by end users continually, distributed over mobile devices, and in many cases considered privacy sensitive. 

To fine-tune models on private, distributed data, 
federated learning is the de-facto approach~\cite{lin2021fednlp,bonawitz2019towards}.
In a training session targeting a specific NLP task and domain, a cloud service selects multiple mobile devices to participate in training. 
A device trains a local copy of the model with its private data and sends the model updates to the cloud. 
Having aggregated model updates from multiple devices, the cloud sends an updated model to the devices. 
The training procedure repeats many rounds (typically hundreds or thousands~\cite{xu2020client, stremmel2021pretraining, mcmahan2017communication}) until the model accuracy reaches a desired level.

As such, this paper focuses on NLP model fine-tuning in a federated setting, a core NLP process in mobile computing. 
Such a process is often referred to as FedNLP~\cite{lin2021fednlp}, for which $\S$\ref{sec:system-model} will present a detailed system model. 


\paragraph{FedNLP overhead} 
Despite the established FedNLP algorithm~\cite{lin2021fednlp}, 
it was unclear if FedNLP is practical on today's mobile platforms.
This paper's first contribution is a thorough characterization of FedNLP on a suite of benchmarks. 
Our results in $\S$\ref{sec:measurement} show prohibitive overheads in twofold: 
(1) Communication.
In each round, participating devices upload local gradients and then download the updated model, 
each transferring hundreds of MBs of data. 
(2) Mobile computation.
Even on an mobile device with GPU, its local computation takes up to several hundred seconds per round. 
As a result, a fine-tuning session can take as long as a few days. 
While federated learning has been known for high overhead in general, 
FedNLP is particularly expensive, primarily because of the large sizes of transformer-based models and NLP task complexity. 
As a comparison, FedNLP's delay is at least 
one order of magnitude
higher than typical federated training delays reported in literature.

\paragraph{Adapters and their configuration}
Our primary goal is to reduce FedNLP's training delay to reach a target accuracy, i.e. time to accuracy.
We first identify \textit{adapters}~\cite{houlsby2019parameter} as key building blocks for NLP models.
As small modules injected between adjacent transformer layers, adapters become the only tunable modules in a pre-trained model, freezing the remaining model parameters (often $>$99\%) which therefore incur no communication or compute overhead. 
While adapters have been proposed for parameter-efficient learning in general~\cite{houlsby2019parameter},
we are the first to identify their significance for FedNLP and investigate the system implications. 

Although adapters significantly reduce tunable parameters and hence the overhead, 
they do not automatically result in optimal training delays. 
The challenge is a large configuration space of adapters: 
to which layer the adapters are injected (depth) and the capacities of individual adapters (width). 
The adapter configuration has a strong impact on training overhead as well as the model convergence delay.
For instance, adding fixed-size adapters to all layers could see up to 2.29$\times$ longer training delay as compared to adding the same adapters to fewer hand-picked layers ($\S$\ref{sec:design:challenges}).
There is no silver-bullet configuration that results in the fastest convergence under each condition;
rather, the optimal configuration depends on
the specific NLP tasks, the target accuracy, and mobile resources such as network bandwidth and local execution speed.
The choice is also dynamic:
even within the same training session, the favorable configuration drifts over time, depending on the model's learning progress.
Picking a non-optimal configuration could slow down model convergence by up to 4.7x and even underperform training with no adapters at all.

\paragraph{Our system: \sys{}}
We therefore present a system called \sys{},
which speeds up FedNLP with two key designs.

First, \sys{} augments the cloud controller with dynamic adapter configuration. 
The key ideas are twofold. 
(1) \textit{Progressive training}.
\sys launches a training session with only small adapters inserted at the model's top layers (i.e. close to the model output), which essentially learns shallow knowledge at a low training cost. 
Only as the model accuracy starts to plateau, \sys adds bottom-layer adapters to training and increases their widths, 
which learns deep knowledge at increasingly higher training costs. 
This resonates with how humans learn knowledge in an incremental fashion and modern learning theories~\cite{bengio2009curriculum}.
(2) \textit{Sideline trials}.
In addition to training the model with a current configuration, 
\sys{} probes \textit{what} the next configuration will be and \textit{when} to switch to it, 
for which \sys{} continuously profiles multiple candidate configurations. 
To do so, 
from all the participating devices in each round, \sys{} allocates multiple trial groups, requests them to train the model with different adapter configurations, and compares their learning progresses. 
If a trial group shows a much higher convergence rate than others as well as the current configuration under training, \sys{} will commit to the configuration of this group. 
While configuration trial distracts some devices from training with the current configuration,
it actually speeds up model convergence. 
This is because the model convergence rate often sees diminishing returns as the population of participant devices grows~\cite{mcmahan2017communication}, 
and our insight is that the surplus devices can better benefit model convergence by profiling next configurations. 


Second, \sys{} enhances devices with cross-round activation cache. 
We exploit an observation: by design, a device trains the same set of adapters in repeated runs until the configuration switches; 
this set of adapters always spans a continuous range of top layers while the bottom layers remain frozen. 
Exploiting such an opportunity, for a given configuration a device only executes a forward pass \textit{once} through the bottom layers, caches the output activations, 
and reuses the cached output as the input to the top layers which will undergo forward and backward passes. 
Caching thus eschews training for the bottom layers, reducing the total training cost by up to one order of magnitude.


\paragraph{Results}
We implement \sys{} atop FedNLP~\cite{lin2021fednlp}, a popular FL framework. 
We test the resultant implementation on NVIDIA TX2~\cite{tx2}/Nano~\cite{nano} and RaspberryPi 4B~\cite{rpi4b}, three development boards with resources similar to mainstream mobile devices. 
On a diverse set of 4 NLP datasets, \sys reduces the training (model convergence) delay from 31.1--124.3 hours to 0.2--4.5 hours (up to 155.5$\times$ reduction) as against a vanilla fine-tuning approach.
During a training session, \sys reduces the network traffic by 126.7$\times$ and per-device energy consumption by 18.4$\times$ on average.
Compared to more advanced methods such as model quantization~\cite{wu2018error}, layer freezing~\cite{guo2019spottune,lin2021fednlp}, and their combination, \sys still brings 4$\times$--48$\times$ speedup.
Our key designs contribute to the results significantly: 
compared to a hand-picked configuration which requires exhaustive offline search, \sys{}'s online configurator reduces the model convergence delay by 4.6$\times$; 
\sys{}'s caching reduces the delay by 3.3$\times$. 
\sys{} is also resource efficient: it reduces the network traffic by 126.7$\times$ and per-device energy consumption by 18.4$\times$ on average.


\paragraph{Contributions} We have made the following contributions. 
\begin{itemize}[leftmargin=*]
\item We carry out the first FedNLP measurement on actual embedded hardware and demonstrate its slow convergence. 
\item We identify adapters as a building block for FedNLP and the major challenge of adapter configuration. 
\item We design an FL framework \sys{} that automatically configures adapters on the fly for fast training and optimizes for mobile resource usage. 
\item We demonstrate \sys's effectiveness through extensive experiments.
For the first time, \sys{} makes FedNLP practical for commodity mobile devices.
\end{itemize}


\section{Background and Motivations}
\label{sec:bkgnd}



\subsection{NLP Training Workflow}
The modern NLP training typically consists of two stages: pre-training and fine-tuning.
During pre-training, a model is trained on large text datasets, e.g., OSCAR corpora~\cite{AbadjiOrtizSuarezRomaryetal.2021} with more than 370 billion words.
Those datasets are obtained from public domains, e.g., Wikipedia, Twitter, etc.
A pre-trained language model captures the linguistic structure that is ubiquitous and independent of downstream tasks.
The pre-training is usually performed in a self-supervised manner and therefore requires no data labels~\cite{erhan2010does}.
It needs huge compute resources (a mid/large GPU cluster)~\cite{brown2020language, devlin2018bert}, typically done by big companies such as Google. 

The fine-tuning adapts the pre-trained model to various, concrete ``downstream'' language tasks such as text classification, sequence tagging, text generation, and question answering.
This often entails modifying the whole or only top layers of the pre-trained model, and additional training passes to adjust the model weights. 
Fine-tuning requires \textit{labeled} samples for the given task, and is done in a supervised fashion.
The downstream tasks are abundant and keep emerging with time, e.g., new domains, topics or data distributions.


The state-of-the-art NLP models that follow the pre-training workflow are transformer-based, e.g., BERT~\cite{devlin2018bert} and its variants~\cite{sanh2019distilbert, hou2020dynabert, liu2020fastbert, sun2020mobilebert, zafrir2019q8bert, bai2020binarybert}.
Those models are composed of many transformer blocks, where each block extensively uses attention mechanisms~\cite{vaswani2017attention}.
We refer readers to recent surveys~\cite{khadhraoui2022survey, han2022survey, han2021transformer} for how those models work internally.
This work specifically targets the fine-tuning stage of transformer-based models for its pivotal role in modern NLP services.

\subsection{Federated Learning}\label{sec:system-model}
The fine-tuning is often performed on data generated by applications in user devices, e.g., input methods~\cite{xu2018deeptype}, emails~\cite{lee2020catbert, sun2019fine}, and instant messaging~\cite{soldevilla2021natural}.
Those data is private by nature and cannot be collected arbitrarily to respect user's privacy concern and legal regulation like GDPR~\cite{voigt2017eu}.
Federated learning (FL)~\cite{kairouz2021advances} addresses this need by enabling many devices to collaboratively train a shared model without giving away their data.
The key idea is to decentralize the training over devices and only ask them to share model updates instead of raw data.
For this reason, the benchmark for NLP training in a federated setting is emerging~\cite{lin2021fednlp}.

\paragraph{System model}
A pre-trained transformer-based language model is given as input.
After that, our task is to fine-tune the model in a federated environment for an unbounded number of unforeseen tasks may emerge. 
For each task, the fine-tuning initiator (or \textit{developer}) specifies how the last output layer shall be revised (e.g., number of classes).

The fine-tuning mostly follows an FL common practice~\cite{bonawitz2019towards}.
In each round, a cloud service (or \textit{aggregator}) selects a fixed number of mobile devices (or \textit{clients}) as participants.
The model is fine-tuned on each client and the model updates will be uploaded/aggregated on the cloud.
The aggregation is often lightweight and the cloud resources can be flexibly scaled out, so its time cost could be neglected.
The fine-tuning speed is mainly bottlenecked by the on-device training and the network transmission.
The most likely network for FedNLP is WiFi, which is highly unstable and constrained from a hundred Kbps to a few Mbps~\cite{xu2021cloud}.
Note that FL typically executes in synchronous manner, so the more constrained hardware/network often bottlenecks~\cite{reisizadeh2020straggler}.
We do not consider the device memory to be an obstacle because:
(1) Modern mobile devices with many GBs of DRAM can support fine-tuning tasks for BERT (BS=8) according to our experiments~($\S$\ref{sec:eval-cost});
(2) Memory inefficiency can be compensated with acceptable training overhead through advanced memory optimizations such as rematerialization~\cite{chen2016training} and paging~\cite{peng2020capuchin}.



The key metric concerned in this work is time-to-accuracy, a widely adopted metric~\cite{coleman2019analysis} that indicates the training time taken to reach a target accuracy.
This is more practical and comprehensive than a single ``time to full convergence'' because the accuracy improvement per time slice dramatically decreases when the model approaches full convergence.
For instance, when FL fine-tuning on task \texttt{AGNEWS}, it takes only 5.2 hours to obtain 80\% accuracy but another 25.9 hours to 90\%.
In our system model, the target accuracy could be preset by the developer so the cloud service will train multiple rounds using FL until the target is met.

We impose no constraint on client selection~\cite{lipyramidfl, nishio2019client, xu2020client, wang2021device, lai2020oort, zhao2021quality, li2021hermes} or training data sampling~\cite{wang2021device, li2021sample} strategies, making it compatible with a mass of recent FL system literature.



	\subsection{Preliminary Measurements}\label{sec:measurement}
\begin{figure*}[t]
	\centering
	\begin{minipage}[b]{0.23\textwidth}
		\hspace*{-10pt}\includegraphics[width=1\textwidth]{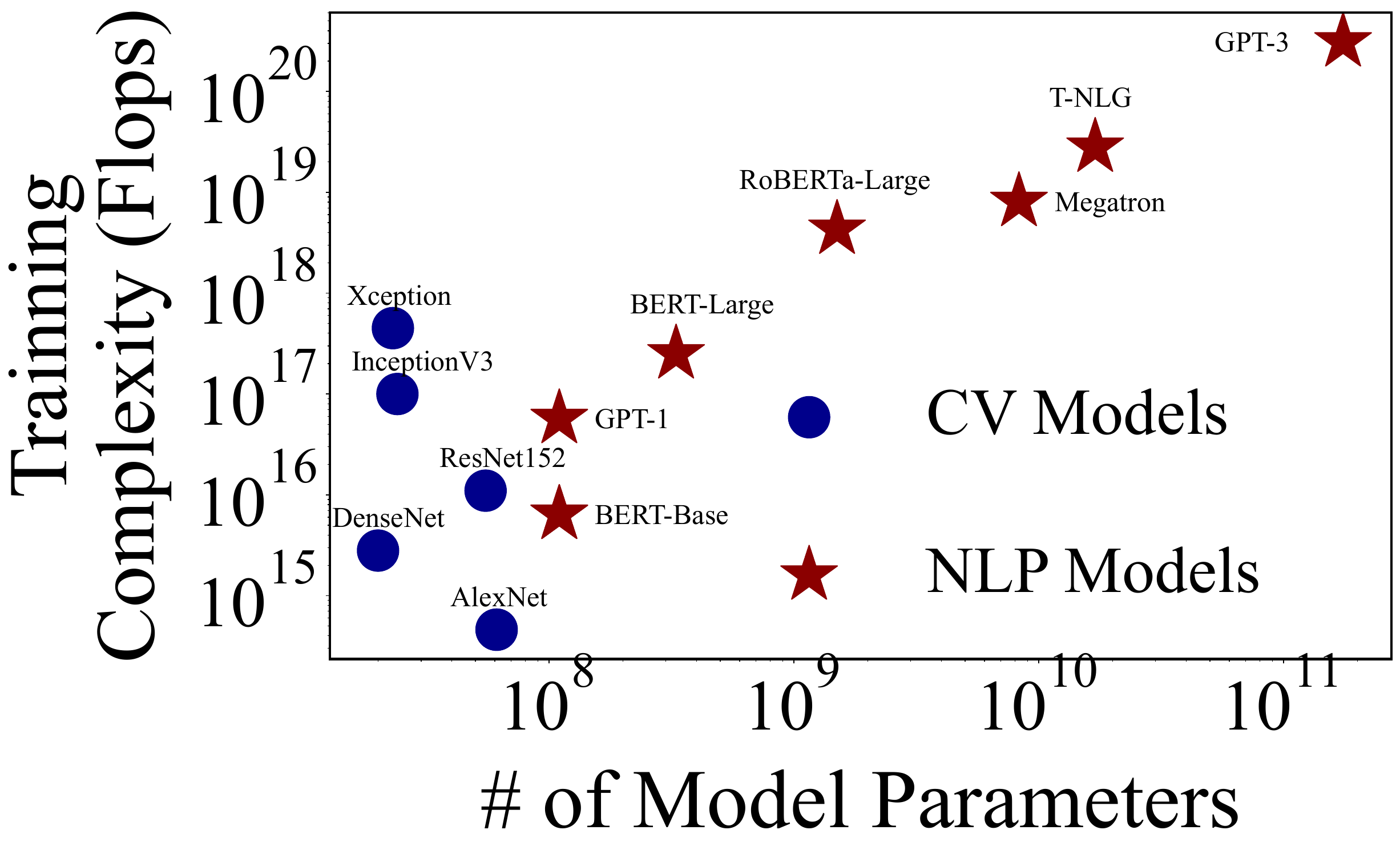}\vspace*{-4pt}
		\subcaption{CV vs. NLP models}
		\label{fig:motivation-cost}
	\end{minipage}
	\hspace*{-3pt}\begin{minipage}[b]{0.23\textwidth}		\hspace*{-5pt}\includegraphics[width=1\textwidth]{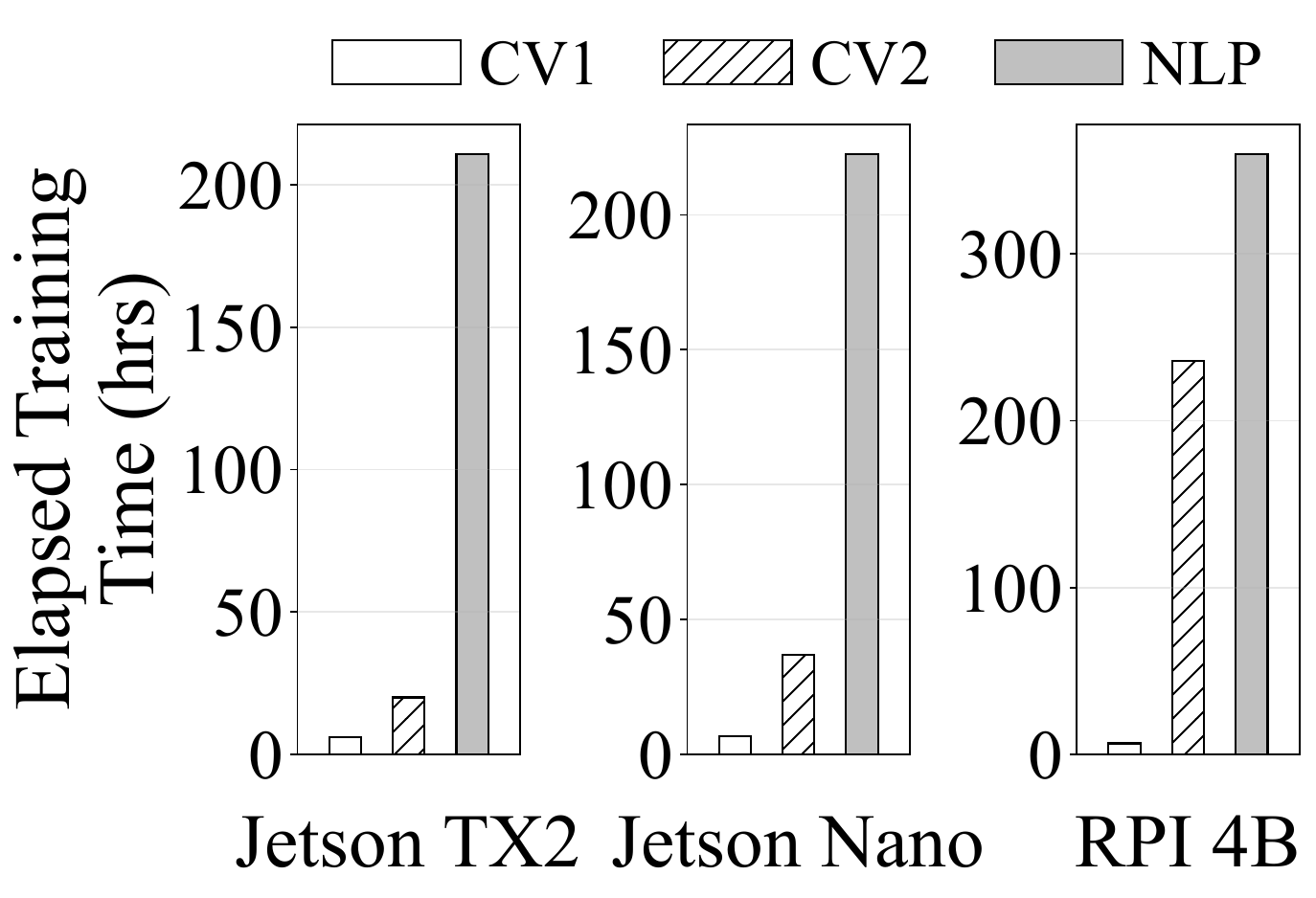}\vspace*{-4pt}
		\subcaption{FL convegence time}
		\label{fig:motivation-convergence}
	\end{minipage}
	\centering
	\hspace*{3pt}\begin{minipage}[b]{0.23\textwidth}
		\hspace*{-10pt}\includegraphics[width=1.1\textwidth]{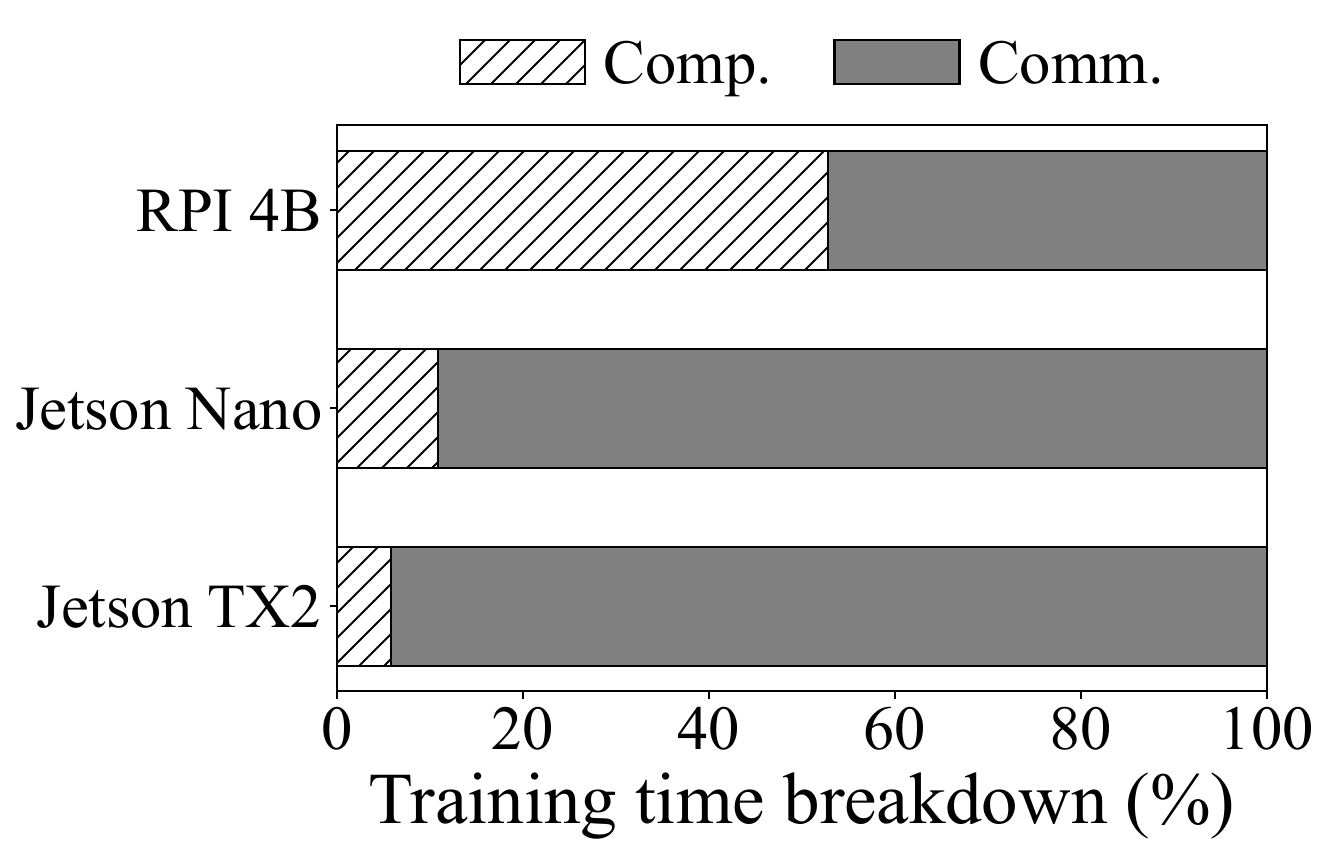}\vspace*{-5pt}
		\subcaption{Breakdown of FedNLP}
		\label{fig:motivation-breakdown}
	\end{minipage}
	\hspace*{8pt}
	\begin{minipage}[b]{0.23\textwidth}
		\hspace*{-7pt}\vspace*{-2.5pt}\includegraphics[width=1.1\textwidth]{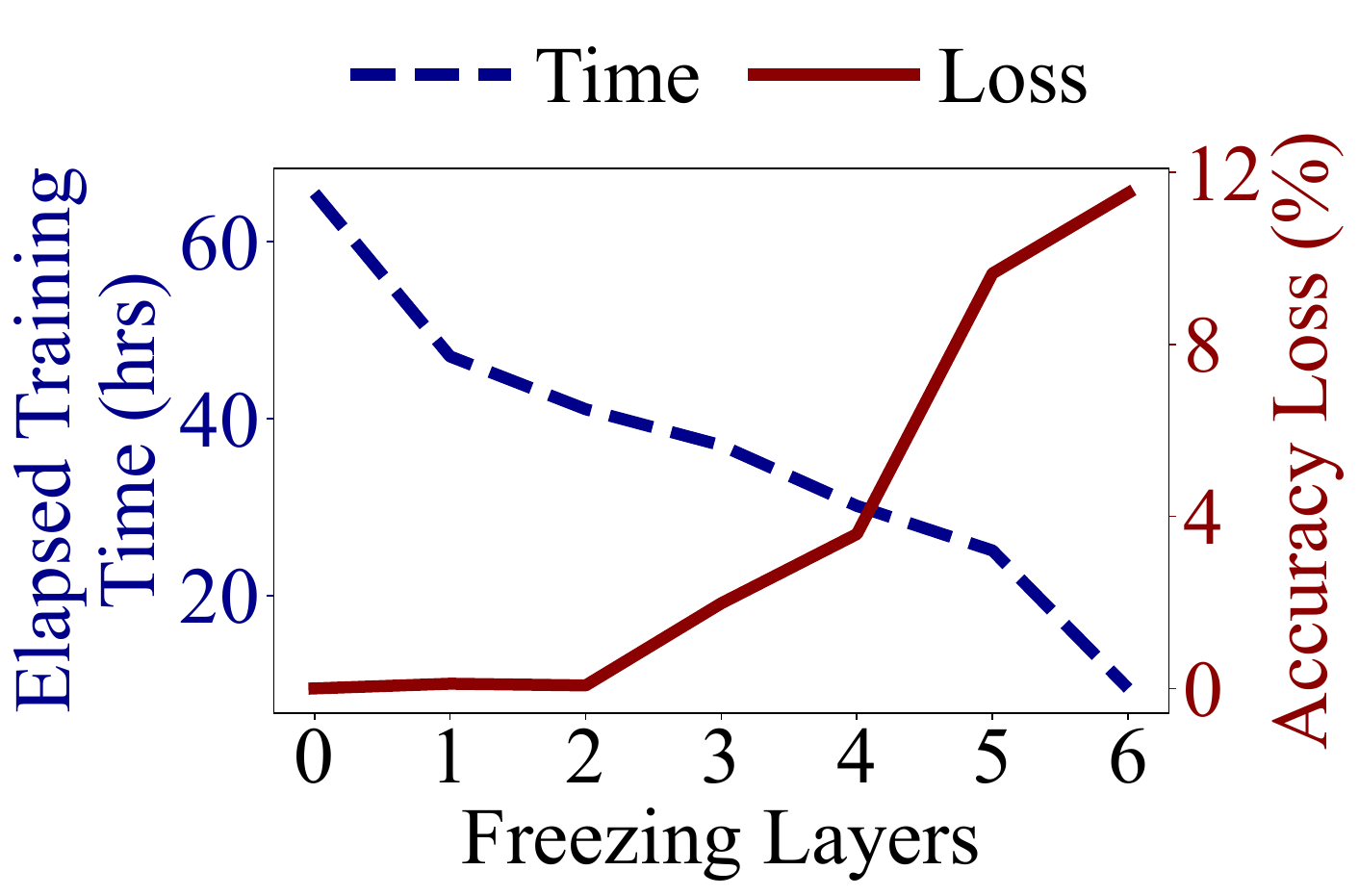}\vspace*{-3pt}
		\subcaption{Layer freezing in FedNLP}
		\label{fig:motivation-freezing}
	\end{minipage}

	\caption{The preliminary measurement results of FedNLP.
	(a) A glance at the complexity of NLP models and traditional CNNs;
	(b) End-to-end convergence time of  CV and NLP models under FL settings (CV1: ``Densenet-121~\cite{huang2017densely} + CelebA~\cite{liu2015deep}''; CV2: ``Resnet56~\cite{he2016deep} + Cifar-100~\cite{krizhevsky2009learning}''; NLP: ``BERT~\cite{devlin2018bert} + Semeval~\cite{hendrickx2019semeval}'').
	(c) Training time breakdown of FedNLP tasks on different hardware. Model: BERT; batch size: 4.
	(d) The performance of layer freezing. Model: DistilBERT~\cite{sanh2019distilbert}; Dataset: ONTONOTES~\cite{pradhan2013towards}; batch size: 4.}
\end{figure*}

We perform preliminary experiments that highlight the motivations to improve FedNLP fine-tuning performance and provide implications for the design of \sys.

\textbf{Observation-1: Transformer-based NLP models are highly costly.}
As illustrated in Figure~\ref{fig:motivation-cost}, transformer-based NLP models are much more expensive than the classic vision models in general in consideration of parameter numbers and computing complexity.
BERT large has 330M trainable weights and takes 250,000 PFLOPs to train, which is 6$\times$/23$\times$ higher than ResNet-152, respectively.

\textbf{Observation-2: FedNLP task is extremely slow.}
Figure~\ref{fig:motivation-convergence} shows the end-to-end training time towards full convergence for typical NLP and CV tasks under federated setting.
As observed, it takes up to 210.74--359.7 hours to train on \texttt{SEMEVAL} dataset, which is 1.53--10.53$\times$ longer than training ResNet-56 on CIFAR-100.
Note that dataset \texttt{SEMEVAL} used for classification tasks has 19 labels, while \texttt{CIFAR-100} has 100 labels.
It's also worth mentioning that the above CV tasks are launched from scratch while the NLP tasks are fine-tuned atop a well pre-trained model.

\textbf{Observation-3: network transmission dominates the training delay on high-end devices.}
The training time spent towards model convergence is dominated by two parts: on-device training and network transmission.
Figure~\ref{fig:motivation-breakdown} shows such breakdown on three kinds of hardware (Jetson TX2~\cite{tx2}, Jetson Nano~\cite{nano}, and Raspberry Pi 4B~\cite{rpi4b}) that span a wide spectrum of hardware capacity and 1MB/s network bandwidth (both uplink and downlink).
It shows that for a high-end edge device like Jetson TX2, the network transmission delay is the major bottleneck (about 94.22\%) of FedNLP tasks.
On a relatively wimpy device, both two parts contribute nontrivially to the total training time.



\textbf{Observation-4: existing techniques are inadequate for FedNLP.}
A common approach to reducing the training cost in NLP fine-tuning is freezing a few bottom transformer blocks~\cite{guo2019spottune,lin2021fednlp}.
It literally reduces the training computations by early stopping the backward propagation and the network cost by only sending the trainable parameters.
Figure~\ref{fig:motivation-freezing} shows the tradeoffs between the convergence time and training accuracy loss by tuning the number of frozen transformer blocks on DistilBERT and \texttt{ONTONOTES}.
Unfortunately, we observed that the profits of such an approach are modest.
For instance, to guarantee an acceptable accuracy loss (e.g., $\le$1\%), 2 out of 6 transformer layers can be frozen at most and only 33.3\% (2/6) network traffic can be saved.

\textbf{\textit{Implications}}
FedNLP is slow due to the considerable amount of time spent on data transmission and local training.
Simply freezing part of the model brings only modest improvement.
While cellular network capacity keeps upgrading, their costly nature hinders adoption in federated tasks~\cite{cheng20225g}.
To enable practical FedNLP with a tolerable convergence delay (e.g., a few hours), the model structure and training paradigm need to be re-architected.
	\section{Design}\label{sec:design}



\subsection{Plugable Adapters}\label{sec:design:adapter}
\begin{figure}[t]
	\centering
	 \includegraphics[width=0.47\textwidth]{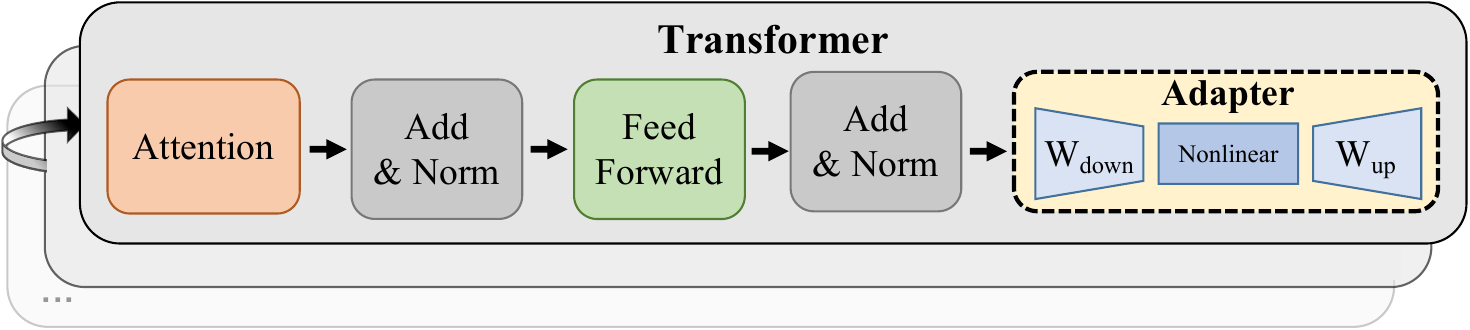}
	\caption{The structure of adapters used.} 
	\label{fig:design-adapter-arch}
\end{figure}



\textbf{Transformer adapters}
For efficient FedNLP, we retrofit adapters -- a recently proposed technique for both CV and NLP tasks to achieve parameter efficiency in machine learning~\cite{houlsby2019parameter}.
The initial goal of adapters is to reduce the tunable parameters especially in continuous learning~\cite{delange2021continual} scenario where unlimited number of new tasks might emerge.
However, it has been seldomly used to tackle system challenges like network cost and convergence speed.
As far as we know, \sys is the first to apply adapters to federated NLP tasks and demonstrate its efficiency under a system context.

The key idea of adapter is to freeze the whole original model but insert a few small modules into different locations inside it.
Figure~\ref{fig:design-adapter-arch} shows the architecture of our adapters and how it's applied to the transformer.
The adapter approach inserts small modules (adapters) between transformer layers. 
The adapter layer generally uses a down-projection with $\boldsymbol{W}_{\text {down }} \in \mathbb{R}^{n \times m}$ to project the input $h$ to a lower-dimensional space specified by bottleneck dimension $m$, followed by a nonlinear activation function $f(\cdot)$, and a up-projection with $\boldsymbol{W}_{\text {up }} \in \mathbb{R}^{m \times n}$. 
These adapters are surrounded by a residual connection, leading to a final form as:
$$
\boldsymbol{h} \leftarrow \boldsymbol{h}+f\left(\boldsymbol{h} \boldsymbol{W}_{\text {down }}\right) \boldsymbol{W}_{\text {up }} .
$$

We follow prior work~\cite{pfeiffer2021adapterfusion}, a state-of-the-art adapter variant to only insert one adapter module after the second sub-layer, i.e., the feed-forward network "add \& layer norm" sublayer.
The output of the adapter is then passed directly into the next transformer layer.

\textbf{The rationales behind adapters}
Why is adapter able to achieve comparable accuracy with much fewer parameters than freezing the bottom transformer layers without revising the model structure?
We reason it with two insights from our experiments\footnote{
    The experiments refer to Figure~\ref{fig:motivation-freezing}, which shows layer freezing exhibits moderated improvement.
}
and related literature~\cite{pfeiffer2021adapterfusion,ruckle2020adapterdrop}. 

First, adapters allow modifying a model's hidden state at a low cost.
By keeping the whole original model as it is, adapters can maximally preserve the knowledge learned from the pre-training dataset. 
The pluggable adapters are only used to encode task-specific representations in intermediate layers of the shared model.
While in fine-tuning scenario, the downstream tasks mostly share low-level feature representation with the pre-training task, it's still beneficial to adjust the low and middle-level feature extractor.
Second, we observe that using adapters stabilizes the convergence process, while fine-tuning on the full model easily goes to overfitting. 
Though the overfitting can be remedied by carefully tuning the hyper-parameters, it also requires non-trivial efforts for each separated fine-tuning task.



\textbf{Network cost analysis}
The trainable parameter number per adapter is $2mn + n + m$.
Clients only send those parameters and last-layer classifier parameters after on-device training to the aggregator.
Therefore the network transmission per round is reduced to 
$D \times (2mn + n + m) + n \times \#labels$,
where $D$ is the total number of transformer blocks of the NLP model. 
As shown in Table~\ref{tab:design-adapter-analysis},
compared to fine-tuning the whole BERT model, the network saving could be more than 99\%.


\textbf{Compute cost analysis}
The computation FLOPs of each adapter in forward pass is $2 \times m \times n \times seqlen$ (normalized to single data sample), where $seqlen$ is the sequence length (default 256 in BERT).
This incurred overhead is trivial compared to the original model complexity, e.g., less than 1\% on BERT.
On the other hand, since all other parameters are fixed during training, calculating the gradients of those fixed weights can be avoided in backward propagation. Table~\ref{tab:design-adapter-analysis} shows adapter brings around 40\% training time reduction.

\begin{table}[t]
	\footnotesize
    \begin{tabular}{c|c|r|r}
    \hline
    \textbf{Model} & \textbf{Method} & \begin{tabular}[c]{@{}c@{}}\textbf{Training Time} \end{tabular} & \begin{tabular}[c]{@{}c@{}}\textbf{Updated Paras.} \end{tabular} \\ \hline
    \multirow{2}{*}{BERT}       & Full Fine-tuning & 1.86 sec & 110.01 x $10^6$ \\ \cline{2-4} 
                                & Adapter & 1.14 sec & 0.61 x $10^6$   \\ \hline
    \multirow{2}{*}{DistilBERT} & Full Fine-tuning & 0.91 sec & 67 x $10^6$     \\ \cline{2-4} 
                                & Adapter & 0.56 sec & 0.32 x $10^6$   \\ \hline
    \end{tabular}
    \caption{Computation and communication cost of inserting adapters into each transformer block (width=32) and full-model tuning on Jetson TX2.}
    \label{tab:design-adapter-analysis}
    \end{table}





\subsection{The Configuration Challenge}
\label{sec:design:challenges}

A unique challenge raised by adapters is its sensitivity to the configurations (explained below).
Different configurations result in a variety of convergence delays, up to 4.7$\times$ gap. 
Choosing an ``optimal'' configuration towards fast convergence is fundamentally challenging for the following reasons.

\paragraph{Large adapter configuration space}
There are two critical parameters of adapters to be determined: depth and width.
(1) Similar to the idea of layer freezing, adapters are not necessarily inserted into each transformer block.
Reducing the number of adapters inserted into the top blocks (namely \textit{tuning depth}) can effectively reduce the network cost and on-device training time.
(2) Apart from the depth, the bottleneck size (\textit{tuning width}), i.e., the target projection dimension of input needs to be carefully set as well.
A small width might not suffice to encode the latent features for fine-tuning tasks and thus incurs high accuracy degradation.
Yet, a too wide adapter incurs high resource costs and therefore slows down the training (i.e., increased time to accuracy).
Overall, the candidate depth spans from 0 to the number of transformer layers even if we only consider inserting adapters at top K consecutive layers, i.e., 12 in BERT, and the valid widths range from 8 to 64 according to our experiments\footnote{The experiments refer to Table~\ref{tab:design-adapter}, which shows that the optimal adapter width ranges from 8 to 64 on four datasets.}.
That results in hundreds of different alternative configurations.

Another dimension of design space is that the configuration can be switched across FL rounds during a training session.
$\S$\ref{sec:design:scheduler} elaborates how such switching could be realized with the knowledge learned by the old adapter configuration well preserved.
But within a round, clients better use the same configuration to facilitate the model aggregation. 

\paragraph{Decisions must be online}
Making a good decision offline is difficult without pre-knowledge about the training dataset -- a common setup in fine-tuning scenarios.
Even with the same task, with the data distribution drifting over time, the resultant model structure could differ tremendously~\cite{li2021hermes,hou2020dynabert}.


\begin{table}[t]
	\footnotesize
	\centering
    \begin{tabular}{l|l|lllll}
    \hline
    \begin{tabular}[c]{@{}l@{}}\end{tabular}
      \multirow{2}{*}{Model} & \multirow{2}{*}{Datasets} & \multicolumn{5}{L{5cm}}{\hspace{-10pt}Optimal adapter configuration \textbf{(depth, width)} 

      \hspace{10pt}towards different target accuracy} \\ \cline{3-7}
      \multirow{4}{*}{\textbf{BERT}} & &
      \multicolumn{1}{l|}{\textbf{99\%}} &
      \multicolumn{1}{l|}{\textbf{95\%}} &
      \multicolumn{1}{l|}{\textbf{90\%}} &
      \multicolumn{1}{l|}{\textbf{80\%}} &
      \textbf{70\%} \\ \cline{1-7}
    & \textbf{20news}  & \multicolumn{1}{l|}{(2,64)}  & \multicolumn{1}{l|}{(2,32)} & \multicolumn{1}{l|}{(2,8)} & \multicolumn{1}{l|}{(2,8)}  & (2,8) \\ \cline{2-7}
    & \textbf{agnews} &
      \multicolumn{1}{l|}{(3,16)} &
      \multicolumn{1}{l|}{(2,16)} &
      \multicolumn{1}{l|}{(2,8)} &
      \multicolumn{1}{l|}{(0,8)} &
      (0,8) \\ \cline{2-7}
    & \textbf{semeval} & \multicolumn{1}{l|}{(10,8)} & \multicolumn{1}{l|}{(6,8)} & \multicolumn{1}{l|}{(6,8)} & \multicolumn{1}{l|}{(2,8)} & (2,8) \\ \cline{2-7}
    & \textbf{ontonotes} &
      \multicolumn{1}{l|}{(12, 32)} &
      \multicolumn{1}{l|}{(12, 32)} &
      \multicolumn{1}{l|}{(10, 32)} &
      \multicolumn{1}{l|}{(0, 16)} &
      (0, 16) \\ \hline
    \end{tabular}
    \caption{The optimal adapter configuration (i.e., best time-to-accuracy) for different target accuracy (ratio to the full convergence) and different datasets.
    }
    \label{tab:design-adapter}
    \end{table}


\paragraph{No silver bullet configuration}
A key observation we made from extensive experiments is that \textit{there is no silver-bullet configuration for FedNLP tasks}.
Rather, the optimal configuration depends on many factors: 
the specific NLP tasks, the target accuracy, and client resources such as network bandwidth and local execution speed.
Even with a given pre-trained model, there are many factors that affect which configuration shall be picked for the fastest convergence.

\begin{figure}[t]
    \centering
    \begin{minipage}[b]{0.23\textwidth}
        \includegraphics[width=2\textwidth]{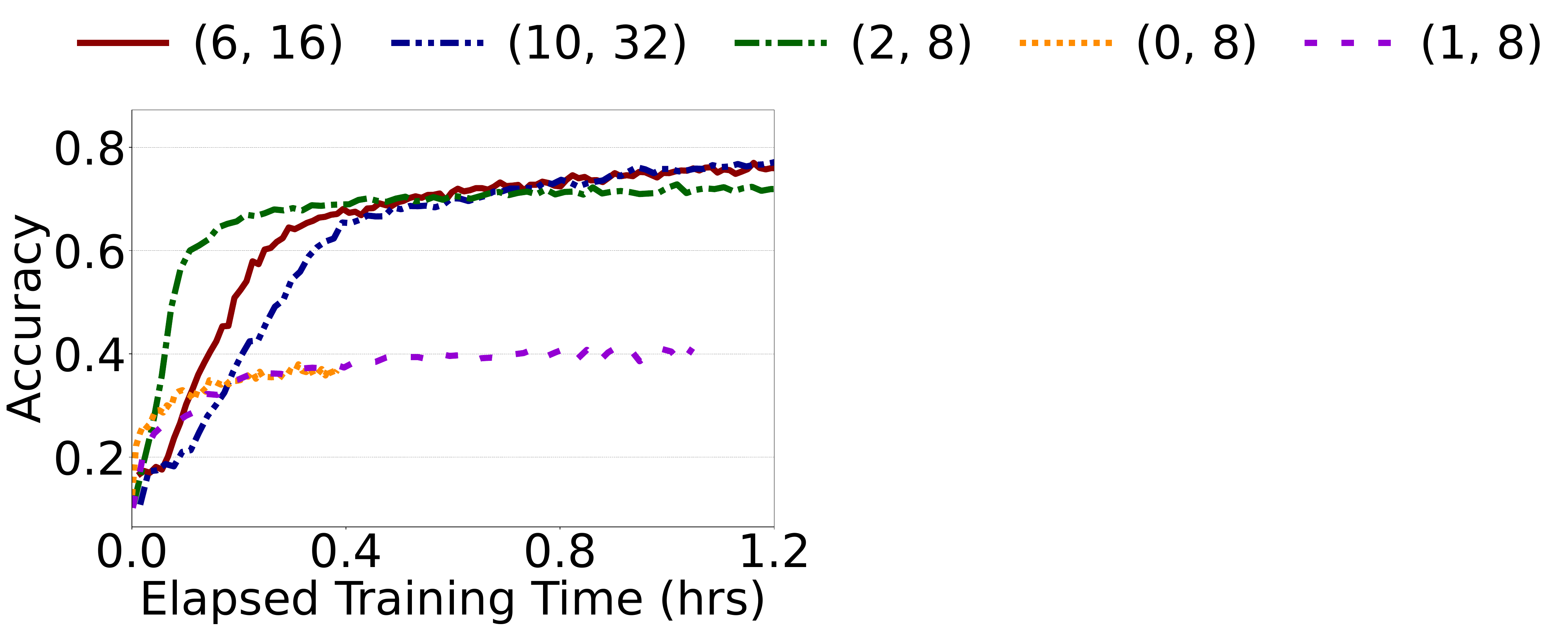}
        \subcaption{\texttt{SEMEVAL}}
    \end{minipage}
    ~
    \begin{minipage}[b]{0.23\textwidth}
        \includegraphics[width=0.98\textwidth]{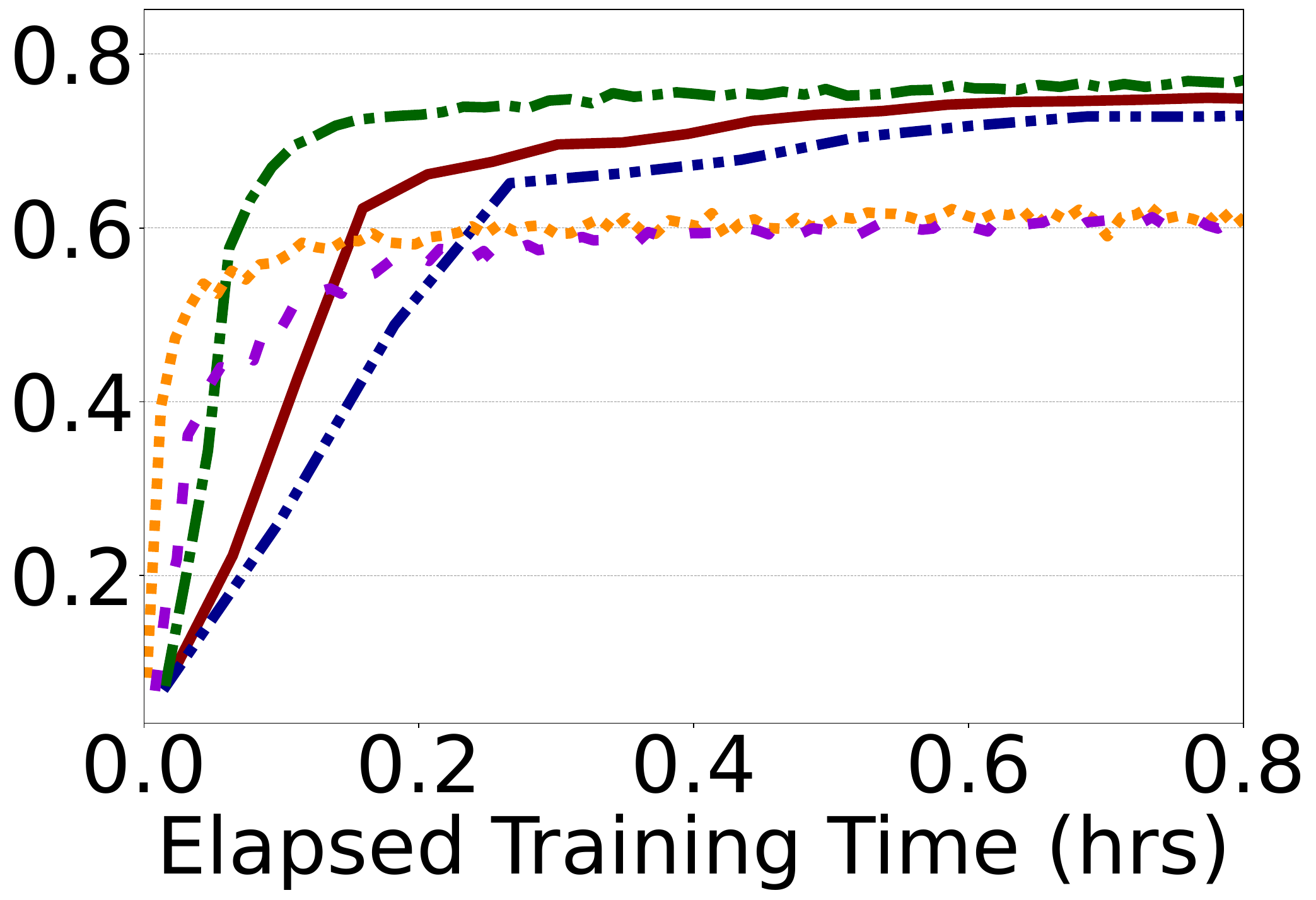}
        \subcaption{\texttt{20NEWS}}
    \end{minipage}
    \caption{Across different target accuracy and  FedNLP tasks, the optimal adapter configuration (depth, width) varies. Tested with BERT and Jetson TX2.}
    \label{fig:design-adapter-configuration}
\end{figure}

\noindent $\bullet$ \textit{Targeted accuracy.}
Within a training session, different target accuracy favors different configurations.
As shown in Figure \ref{fig:design-adapter-configuration} (a), to achieve the best accuracy possible, using an adapter with depth 6 and width 16 is the best option.
If 80\% relative accuracy is satisfactory, the adapter with depth 2 and width 8 is 2$\times$ faster than the previous configuration.

\noindent $\bullet$ \textit{Targeted NLP tasks.}
Across different FedNLP tasks, the optimal adapter configuration varies.
As shown in Figure \ref{fig:design-adapter-configuration}, using depth 2 and width 8 leads to fast convergence to 80\% accuracy on \texttt{20NEWS}~\cite{lang1995newsweeder}. However, on \texttt{SEMEVAL}~\cite{hendrickx2019semeval}, the same configuration results in 10\% lower convergence accuracy as compared to a more complex configuration.

\noindent $\bullet$ \textit{Client resources.} 
Local clients' training speed and network capacity also make a difference in adapter selection.
This is due to the disparate impacts of adapter depth/width on the computation and communication reduction.
For instance, a larger tuning depth linearly amplifies the communication and computation cost, while a larger tuning width linearly increases the communication cost but only adds negligible computation cost according to the analysis in $\S$\ref{sec:design:adapter}.


\paragraph{Why prior work is inadequate}
A closely related technique is neural architecture search (NAS)~\cite{elsken2019neural}, which automatically looks for the best model structures but with a totally different design goal.
Essentially, NAS sacrifices the time for good training accuracy, e.g., days to train a single model in a centralized manner~\cite{wei2022npenas}.
Instead, we pursue fast time-to-accuracy, which is a more practical and affordable setting for FedNLP developers.

\subsection{The Online Configurator}
\label{sec:design:scheduler}

We build a configurator that automatically adjusts the tuning depth and width throughout a training session. 
The goal is fast model convergence: achieving the target model accuracy in the shortest time. 
Our key ideas are twofold:

$\bullet$ \textit{Progressive training.}
The first key idea is, to begin with a shallow tuning configuration (i.e., small depth and width) to quickly boost the model accuracy.
When it encounters a ``choke point'' where more rounds of training no longer provide enough accuracy profit, it ``upgrades'' to a more complex configuration, i.e., either deeper or wider.

Such upgrading mechanism is inspired by curriculum learning~\cite{bengio2009curriculum}, a learning strategy that trains a model beginning from easier data samples to harder ones.
Instead of altering the training samples, we propose to alter the model structure.
In the beginning, a simpler adapter configuration can learn fast.
This is because, by focusing on fewer compact trainable parameters closer to the model output, the model can rapidly learn the coarse-grained domain-specific knowledge for the downstream tasks, such as new class labels~\cite{ro2021autolr}.
For simple downstream tasks, fine-tuning without re-learning deep features is enough to obtain satisfactory model accuracy, e.g., depth 2 and width 64 for 20NEWS dataset~\cite{lang1995newsweeder}.
As the training proceeds, the model encounters a ``choke point'' where the learning curve becomes gentle.
It demands deeper or wider adapters to learn new features.
The experiment results in Table~\ref{tab:design-adapter} attests to our claim that a higher target accuracy favors deeper and wider adapters.

$\bullet$ \textit{Identifying timing and direction to upgrade configuration through sideline trials.}
The learning curve is fundamentally challenging to be estimated or predicted ahead of time.
How can a system possibly know the timing and to which direction to upgrade?
In this work, we propose an intuitive approach based on the concept of sideline trials.
Its key idea is to ask extra participant clients to attempt different configurations, and make a decision on whether and where to upgrade based on the tested accuracy of different directions.
In federated settings, such ``extra clients'' are common because the client-level parallelism of existing FL algorithms is notoriously low.
That is, limited by the learning theory~\cite{keskar2016large}, a small number of clients (i.e., 5 for \texttt{20NEWS}) is enough to saturate the convergence performance (both accuracy and speed) and allocating more clients gives a negligible return.
As will be shown in $\S$\ref{sec:eval-ablation}, using those extra clients for trial is much more beneficial than asking them to participate in training.

\begin{algorithm}[t]
	\footnotesize
	\SetAlgoNoEnd
  \SetKwProg{Fn}{Function}{~}{end}
  \SetKwData{Left}{left}\SetKwData{This}{this}\SetKwData{Up}{up}
  \SetKwFunction{MIN}{MIN}\SetKwFunction{MAX}{MAX}\SetKwFunction{LENGTH}{LENGTH}
  \SetKwInOut{Input}{input}\SetKwInOut{Output}{output}\SetKwInOut{Variable}{Variable}

  \Input{
      Target accuracy, $acc$;\\
      Trial interval, $trial\_intvl$;\\
      Start-up depth and width, $D_{0}$ and $W_{0}$;\\
      Step of depth and width, $S_d$ and $S_w$.
    }

  \Output{
      Fine-tuned adapter weights, $\Theta_{i}$~(i=1,2\dots).
    }
  \BlankLine

  \Fn{Cloud\_controller():}{
    $Trial_{0}$, $Trial_{1}$, $Trial_{2}$ $\leftarrow$ selects $3N$ clients;\\
    Iteration i=0;\\
    $T_{trial}$=$T_{now}$;\\
    Dispatch(0); // Init model and trigger client training\\
      \While{Eval() $<$ $acc$}
      { 
        i++;\\
        $\Theta^{k}_{i}(n)$ $\leftarrow$ Receive updated adapters from $Trial_{k}$;\\
      $\Theta^{k}_{i}$ $\leftarrow$ Fedavg($\Theta^{k}_{i}(n)$);\\
      $Trial_{0}$, $Trial_{1}$, $Trial_{2}$ $\leftarrow$ selects $3N$ new clients;\\
        \eIf{$T_{now}$ - $T_{trial}$ > $trial\_intvl$}{Compare accuracy under different $\Theta^{k}_{i}$;\\
        $\Theta_{i}$ $\leftarrow$ Winner track;
        $D_{i}, W_{i}$ $\leftarrow$ Winner setting;\\
        $T_{trial}$=$T_{now}$;\\
        Dispatch(i) // Inherit winner track;\\}{
          Send the aggregated model $\Theta^{k}_{i}$ to $Trial_{k}$.
        }
      }
      Exit training.
    }

  \Fn{Client\_training(i,k):}{
    $\Theta_{i}^{k}$ $\leftarrow$ Receive global adapter  from cloud;\\
    $\Theta_{i+1}^{k}(n)$ $\leftarrow$ Train and update local adapter; \\
    Send updated adapter $\Theta_{i+1}^{k}(n)$ to cloud.
  }
  
  \Fn{Dispatch(i):}{
    $F$($D$,$W$): Initial/Inherit winner track $\Theta_{i}$ with $D$, $W$.
  
  $\Theta_{i}^{0}$ $\leftarrow$ $F$($D_{i}$,$W_{i}$);
  $\Theta_{i}^{1}$ $\leftarrow$ $F$($D_{i}+S_{d}$,$W_{i}$);
  $\Theta_{i}^{2}$ $\leftarrow$ $F$($D_{i}$,$W_{i}+S_{w}$);\\
  Sends $\Theta_{i}^{0}$, $\Theta_{i}^{1}$, $\Theta_{i}^{2}$ to $Trial_{0}$, $Trial_{1}$, $Trial_{2}$ seperately\;
  Parallel: Client\_training(i,k).
  }
  
  \caption{Our Online Configurator}\label{alg:trial}
\end{algorithm}

\paragraph{Configurator algorithm in detail}
Algorithm~\ref{alg:trial} shows how \sys progressively upgrades the configuration of adapters during a training session.
Unlike the traditional FL scheme where only one global model with a fixed structure undergoes the training, in \sys the cloud aggregator periodically dispatches the global model to three groups of clients: one is to train with the current configuration, one with a deeper one and the other with a wider one (line 2--5, 23-26).
After a few rounds of parallel training (line 27, 19--22, 7--10, 18), the aggregator server checks the accuracy of three global models and re-starts the process on the model with the highest accuracy (line 12--15).
Note that when the aggregator checks the accuracy, the three global models undergo different numbers of global rounds because the per-round training time and network time depend on the adapter configuration ($\S$\ref{sec:design:adapter}).
Therefore, the training speed of different tuning depth/width is considered in this mechanism.
Except that, the clients and aggregator follow the common FL process in local training (line 19--22) and model aggregation (line 8--9).

As described in Algorithm~\ref{alg:trial} (line 23--27), the models dispatched to different groups are with different model configurations.
Group $Trial_{0}$ inherits the learned adapter from the previous winner track whereas $Trial_{1}$ and $Trial_{2}$ also inherit the old adapters but add extra depth and width, respectively.
The step sizes of depth and width are pre-defined, e.g., $S_{d}=1$ and $S_{w}=8$ by default, respectively.
The depths and widths are both added on the fly.
All newly added weights are normalized with mean (0.0) and stddev (0.02).
We experiment with two ways to expand the adapter width: vertical and parallel stacking.
Our micro experiments show that vertically stacking will outperform parallel stacking with the same amount of parameters.
So we always use this method.

\textbf{Integration with the existing FL frameworks}
\sys{} is compatible with how existing FL frameworks manage clients for training efficiency, 
a key system component having received high research attention~\cite{lipyramidfl, nishio2019client, xu2020client, wang2021device, lai2020oort, zhao2021quality, li2021sample}. 
This is because the adapters and their configuration scheduler are intentionally designed to be decoupled from which device or data will be involved in per-round training.

\textbf{Fairness across trial groups}
\sys could make a fair comparison across trial groups under client/data heterogeneity.
\sys uses a two-stage client selection: it first follows the used FL method to select clients (might be prioritized and biased), and then “randomly” partitions them into different groups.
It statistically ensures a fair comparison across trial groups.
The same notion applies to data heterogeneity as well.
Moreover, \sys re-selects clients/data for each group per round. 
It compares performance across groups every N round.
\sys hence mitigates potential unfairness given a large N (N>10), per the sampling theory.


\textbf{User overhead with sideline trials}
The extra cost incurred by the sideline trials is amortized across different clients and shall not impose higher impacts on user experience. 
In an end-to-end manner, using trial groups reduces the elapsed training time towards convergence by discovering better adapter configurations, thus reducing the total cost accumulated on all clients as reported in $\S$\ref{sec:eval-ablation}. 
This design makes \sys more scalable to available clients: using those extra clients for trial is much more beneficial than asking them to participate in training, either in convergence speed or end-to-end resource cost.


\section{FedNLP Activation Cache}\label{sec:design:cache}




Using adapter ($\S$\ref{sec:design:adapter}) significantly reduces the network cost in FedNLP tasks, exposing client-side computation as the next major bottleneck for model convergence.
Layer-freezing reduces the training computations by early-stopping the backward propagation, but does not address the computation cost at forward pass.
For instance, with tuning depth as 2 of 12 transformer blocks in BERT, the forward computation takes $12/(12+2*2)=75\%$ of the total computation, considering that the backward propagation computation of each layer is approximately 2 times of the forward counterpart~\cite{qi2016paleo}.

\textbf{Opportunities}
To reduce the client computation, we exploit two unique FedNLP opportunities:
(i) Throughout a training session, a device participates in many FL rounds.
Across rounds, the device executes local training on the same set of local samples.
In typical FL scenarios, it takes tens of thousands of rounds till model convergence~\cite{bonawitz2019towards}.
(ii) During on-device training, the weights of the bottom transformer blocks that do not have adapters inserted are fixed.
Moreover, those untouched transformer blocks remain the same across FL rounds till \sys switches the tuning depth ($\S$\ref{sec:design:scheduler}).
In our experiments we observe a tuning depth upgrading interval typically at hundreds to thousands of rounds.

\textbf{The activation cache}
Our key idea is to leverage both the static input and the fixed bottom layers so a client's computed activations can be reused across rounds.
Assuming the cloud aggregator selects a device to train with depth $d_{prev}$ at round $r_{prev}$.
During the training, the device also extracts the output of the last fixed bottom layer for each input batch, i.e., the output of the $(D-d_{prev})_{th}$ transformer layer and stores them in local storage.
For the next time device $k$ is selected, it first inquires the cloud aggregator that how deep the model has been tuned since the last time it's selected, i.e., $d'=\max(d_{prev}, d_{prev+1}, \dots, d_{now})$.
If $d'>d_{prev}$, it means the model has gone deeper.
The transformer layers between $(D-d')_{th}$ and $(D-d_{prev})_{th}$ have been updated and the cached activations of $(D-d_{prev})_{th}$ transformer block have ``expired''.
The output of the $(D-d')_{th}$ transformer layer needs to be recomputed and recached.
Otherwise, if $d'\leq d_{prev}$, the first $d_{split}=D-d_{prev}$ transformer layers are not touched since round $r_{prev}$.
Therefore, it can directly load the output of the $(D-d_{prev})_{th}$ layer from the cache and feed it into the training process without starting from scratch.
The above process repeats when the device participates in training every time.
The cache expiration incurs re-computations of bottom transformer blocks and compromises its profits.
Fortunately, \sys's design of online configurator ($\S$\ref{sec:design:scheduler}) orchestrates with the caching technique by monotonously upgrading the tuning depth.

To be noted, while cache mechanism has been widely exploited to accelerate on-device DNN inference~\cite{crankshaw2017clipper,shen2019nexus,han2016mcdnn}, we are the first to sense its opportunity in training tasks (FedNLP in this case) that naturally involves repeated computations on the same data.


\textbf{Computation and storage cost analysis}
Using activation caching reduces the computations by $d_{split}/D$ at the forward pass.
Yet it also takes extra storage, i.e., $seqlen \times n \times BS$ per batch,
where $n$ is the transformer's internal feature size (default 768), and $BS$ is the batch size (default 4).
The cache is reloaded from disk per minibatch, taking no more than tens of ms on embedded flash and incurs less than 2\% overhead. 
The total cache size is also proportional to the number of batches samples per client (typically dozens).
Assuming 100 training samples, the storage cost is calculated to be around 100MB.
Such cost is no more than 1\% of the storage of a modern mobile/embedded device, e.g., tens to hundreds of GBs.
The cache can be cleared once the FL process finishes.

\textbf{Caching efficiency under millions of clients}
Our activation caching might lose efficiency in the face of millions of available clients, each of which is selected in one or very few rounds. 
But we expect such scenarios as uncommon from two folds.
For one thing, an FL process typically needs thousands of or even more rounds to reach convergence.
For another thing, in practice, FL employs rigorous client selection strategies, e.g., a training-eligible device must be charged, wifi-connected, and battery-sufficient. According to Google~\cite{bonawitz2019towards}, the utmost eligible client number is only 0.1\% (10K out of 10M) at midnight and could even be 10$\times$ smaller at other times. Such observation is also confirmed by another FL literature~\cite{yang2021characterizing} that reports a large portion of devices will never participate in the FL process.
The advanced client selection methods~\cite{nishio2019client, xu2020client, wang2021device, lai2020oort, zhao2021quality} that favor strong devices make the participant devices even more skewed and so those devices will participate in many rounds of training.

	\section{Implementation and Setups}\label{sec:impl}

We have fully implemented the \sys prototype atop \texttt{FedNLP} \cite{lin2021fednlp} (the SOTA framework to evaluate FL methods on NLP tasks) and \texttt{Adapterhub}~\cite{pfeiffer2020adapterhub} (a library that facilitates the integration of pre-trained adapters for different tasks).
As prior work~\cite{bonawitz2019towards}, we adopt the parameter server (PS) architecture among the clients and central server.
At the server side, once job is submitted by the developer, the server initializes the pluggable meta adapter to be trained (through the API of \texttt{Adapterhub}) into the pre-trained model.
The server also splits the initialized meta adapter into three branches: normal, wider and deeper.
The wider branch will stack a few meta adapters parallel to expand the bottleneck size of adapter in single layer.
The deeper branch will insert the meta adapter into one more deeper layer.
A client selector will sample 3N clients from available devices and shuffle them into 3 groups.
We now employ a random client selector (default in most FL literature) but more advanced selection strategies~\cite{li2021hermes, lipyramidfl, nishio2019client, xu2020client, wang2021device, lai2020oort, zhao2021quality, li2021sample} can be plugged into our implementation as well.
Then, the server sends three branches of adapters to three groups separately via MPI (in standalone mode) or WLAN/Cellular (in distributed mode).
Once receiving the adapters, the clients insert the adapter into their local pre-trained model.
They fine-tune the model with their own private data.
The trained adapters will be collected in the central server and aggregated through \textit{FedAvg} algorithm~\cite{mcmahan2017communication}.
All clients run in synchronized mode~\cite{ho2013more}.

\begin{table}[]
    \footnotesize
    \begin{tabular}{l|L{4cm}|L{1.5cm}}
    \hline
    \textbf{Device} &
      \textbf{Processor} &
      \textbf{Per-batch Latency (s)} \\ \hline
      Jetson TX2~\cite{tx2} & 256-core NVIDIA Pascal™ GPU. &
      0.88 \\ \hline
      Jetson Nano~\cite{nano} & 128-core NVIDIA CUDA® GPU. & 1.89 \\ \hline
    RPI 4B~\cite{rpi4b} &
      Broadcom BCM2711B0 quad-core A72 64-bit @ 1.5GHz CPU. &
      18.27 \\ \hline
    \end{tabular}
    \caption{Development boards used in experiments.}
    \label{tab:eval-setups-device}
    \end{table}

\textbf{Metrics}
We mainly report the time-to-accuracy metric.
We divide the dataset of each device for training (80\%) and testing (20\%).
For clarity, we pay attention to a few typical accuracy targets, e.g., 99\%, 95\%, 90\% of the full convergence accuracy achievable by the baseline that fine-tunes the whole model.
We refer to those accuracy numbers as \textit{\relativeacc}.
For example, the 100\% \relativeacc of BERT is 0.8 (accuracy) for \texttt{20NEWS}; 0.9 (accuracy) for \texttt{AGNews}; 0.8 (accuracy) for \texttt{SEMEVAL}; and 0.75 (token-F1) for \texttt{ONTONOTES}.
We also report the resource cost in an FL process, including the total energy consumption on data transmitting and training computation on each client; the total amount of network traffic; and the peak memory usage.

\textbf{Hardware}
As prior FL literature~\cite{lin2021fednlp,li2021hermes,lipyramidfl,lai2020oort,shin2022fedbalancer}, our experiments are carried out in an emulation manner on a GPU server with 8x NVIDIA A40.
The on-device training time is obtained on 3 development boards with similar hardware capacity to mainstream mobile devices, i.e., Jetson TX2~\cite{tx2}, Jetson Nano~\cite{nano}, and Raspberry Pi 4B~\cite{rpi4b}.
The numbers are then plugged into the emulation framework to calculate the elapsed time.
The default network bandwidth between clients and server is set to 1MB/s, a typical setting for mobile and IoT devices~\cite{wifi-state,han2016mp}.
Note that while home/office WiFi downlink could be faster, the uplink bandwidth is often bound by the 
broadband backbone~\cite{huang2013lte}.
In $\S$\ref{sec:eval-e2e}, we will also quantify the performance of \sys under various hardware and bandwidth settings (100KB/s--10MB/s).

\textbf{Models}
We use two representative models for FedNLP tasks: BERT~\cite{devlin2018bert} (default) and its varient DistilBERT~\cite{sanh2019distilbert}.
BERT and DistilBERT are composed of 12 and 6 transformer blocks, respectively.
DistilBERT leverages knowledge distillation during the pre-training phase and reduces the size of a BERT model by 40\%, while retaining 97\% of its language understanding capabilities and being 60\% faster.
We use BERT for most of our experiments, as all BERT-based variants derive from it.
The pre-trained weights of both models are downloaded directly from Hugging Face~\cite{wolf2019huggingface}.

\begin{table}[]
    \footnotesize
    \begin{tabular}{@{}cccccl@{}}
    \toprule
    \textbf{Task} & \textbf{Dataset} & \textbf{\# of Clients} & \textbf{Labels} & \textbf{Non-IID} & \textbf{Samples} \\ \midrule
    TC & \texttt{20NEWS}~\cite{lang1995newsweeder}  & 100               & 20      & /       & 18.8k        \\
    TC & \texttt{AGNEWS}~\cite{zhang2015character}  & 1,000               & 4       & a=10     & 127.6k        \\
    TC & \texttt{SEMEVAL}~\cite{hendrickx2019semeval} & 100               & 19      & a=100   & 10.7k        \\
    ST & \texttt{ONTONOTES}~\cite{pradhan2013towards}    & 600                & 37      & a=10     & 5.5k        \\ \bottomrule
    \end{tabular}
    \caption{Datasets and settings used in experiments for \textbf{T}ext \textbf{C}lassification and \textbf{S}equence \textbf{T}agging. ``a'' is a parameter that controls the datasets' non-IID level~\cite{lin2021fednlp}.}
    \label{tab:datasets}
\end{table}
\textbf{Tasks and datasets}
We evaluate \sys on 4 classic NLP downstream datasets as shown in Table~\ref{tab:datasets}.
We follow the approach in~\cite{lin2021fednlp} to build the non-IID datasets.
(1) \texttt{20NEWS} (IID)~\cite{lang1995newsweeder} dataset is a collection of approximately 20,000 newsgroup documents.
(2) \texttt{AGNEWS} (non-IID)~\cite{zhang2015character} is a collection of 127.6K news articles gathered from more than 2,000 news sources.
(3) \texttt{SEMEVAL} (non-IID)~\cite{hendrickx2019semeval} is a relation classification datasets which assigns predefined relation labels to the entity pairs that occur in texts.
The above 3 datasets are used for text classification (TC)~\cite{sun2019fine} tasks, where the output is a label in a fixed set of label set (e.g., political, sports, and entertainment).
(4) \texttt{ONTONOTES} (non-IID)~\cite{pradhan2013towards} is a corpus where sentences have annotations for the entity spans and types.
This dataset is for sequence tagging (ST)~\cite{aras2021evaluation} task, where the output is a sequence of tags.

\begin{table*}[ht]
  \begin{tabular}{c|ccc|ccc|ccc|ccc}
  \hline
  {\textbf{Datasets}} & \multicolumn{3}{c|}{\textbf{\texttt{20NEWS}}} & \multicolumn{3}{c|}{\textbf{\texttt{AGNEWS}}} & \multicolumn{3}{c|}{\textbf{\texttt{SEMEVAL}}} & \multicolumn{3}{c}{\textbf{\texttt{ONTONOTES}}} \\
           Relative Accuracy & 99\% & 95\% & 90\% & 99\% & 95\% & 90\% & 99\%  & 95\% & 90\% & 99\% & 95\% & 90\% \\ \hline
           \texttt{FT}  & 44.0 & 23.4 & 13.1 & 31.1 & 10.1 & 5.2  & 124.3 & 89.9 & 61.7 & 76.1 & 55.9 & 35.6 \\ \hline
           \texttt{FTQ} & 12.7 & 6.8  & 3.8  & 9.1  & 2.6  & 1.7  & 32.0  & 23.1 & 15.9 & 21.2 & 15.5 & 9.9  \\ \hline
           \texttt{LF$_{oracle}$}   & 18.5 & 8.1  & 4.3  & 9.6  & 1.4  & 1.1  & 74.0  & 46.8 & 33.2 & 82.5 & 43.8 & 24.5 \\ \hline
           \texttt{LFQ$_{oracle}$} & 5.2  & 2.5  & 1.1  & 1.6  & 0.3  & 0.2  & 16.8  & 11.0 & 7.7  & 23.9 & 12.9 & 7.2  \\ \hline
  \textbf{\texttt{\sys}}     & \textbf{1.3}  & \textbf{0.4}  & \textbf{0.1}  & \textbf{0.2}  & \textbf{0.03} & \textbf{0.02} & \textbf{2.3}   & \textbf{1.1}  & \textbf{0.6}  & \textbf{4.5}  & \textbf{2.4}  & \textbf{1.3}  \\ \hline
  \end{tabular}
  \caption{Elapsed training time taken to reach different \relativeacc. NLP model: BERT. Unit: Hour.
	}    
  \label{tab:runtime-performance}
  \end{table*} 
\begin{figure*}[t]
    \centering
    \begin{minipage}[b]{1\textwidth}
        \begin{minipage}[b]{0.24\textwidth}
            \includegraphics[width=1.0\textwidth]{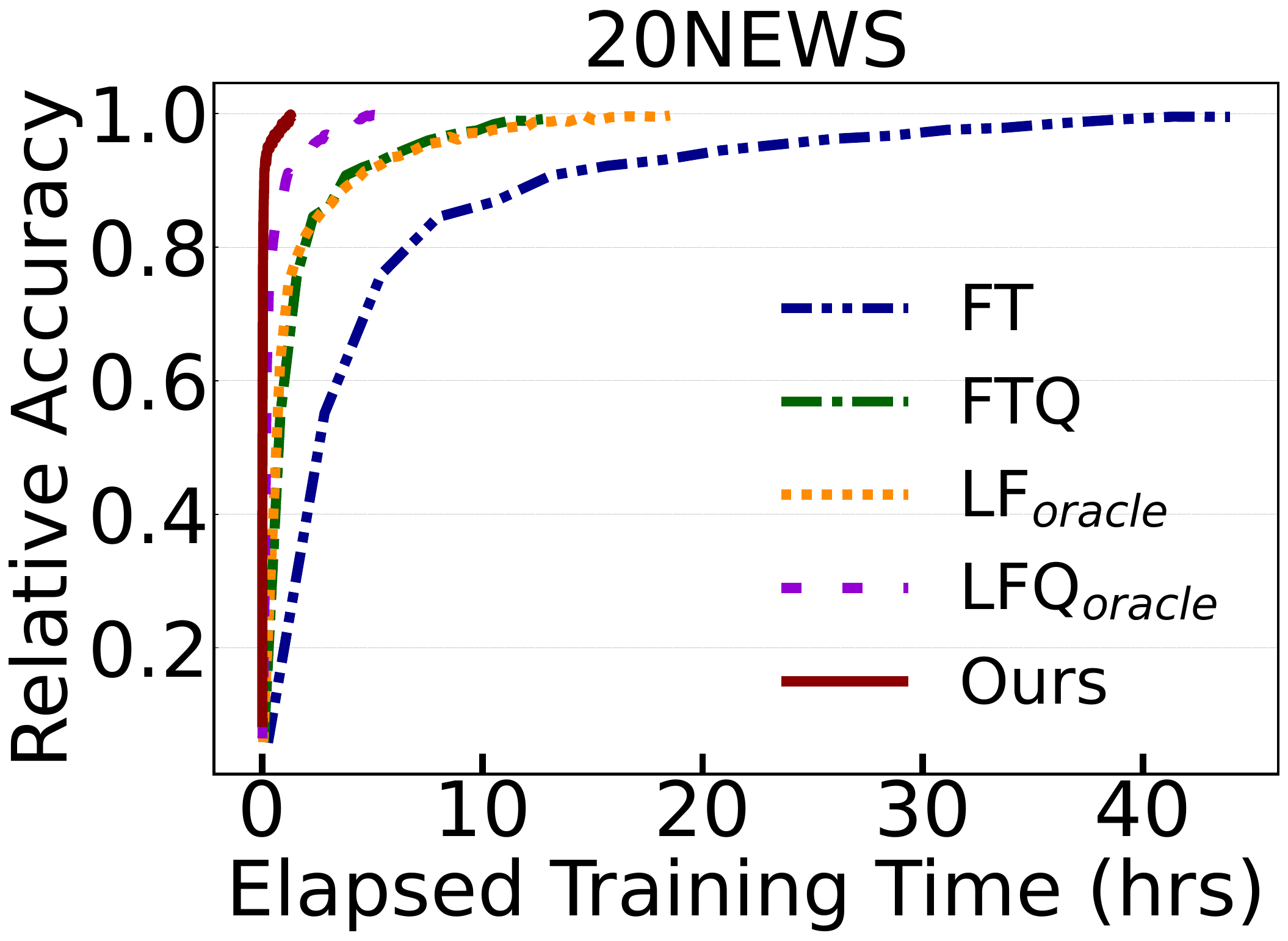}
        \end{minipage}
        ~
        \begin{minipage}[b]{0.24\textwidth}
            \includegraphics[width=0.95\textwidth]{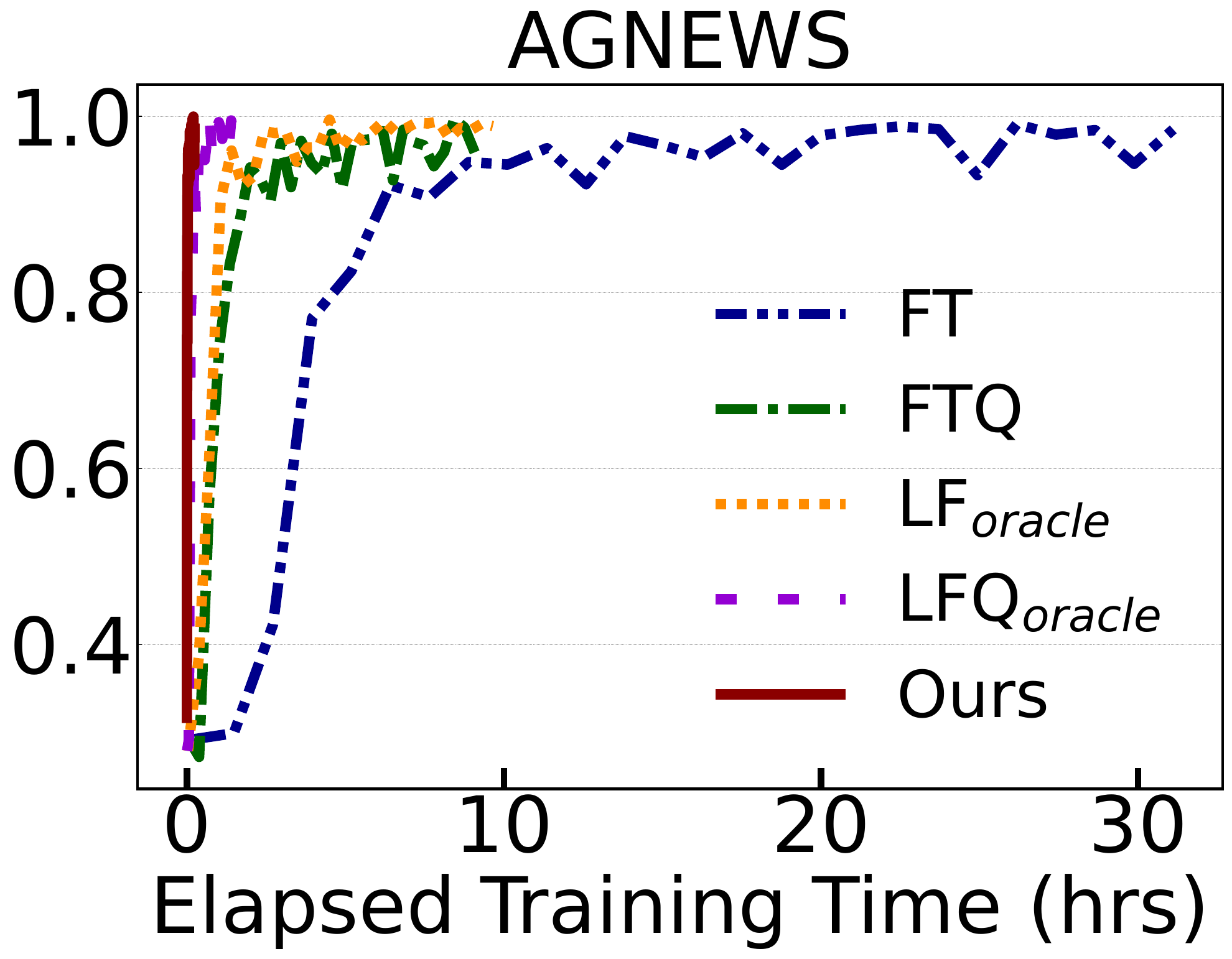}
        \end{minipage}
        ~
        \begin{minipage}[b]{0.24\textwidth}
            \includegraphics[width=0.95\textwidth]{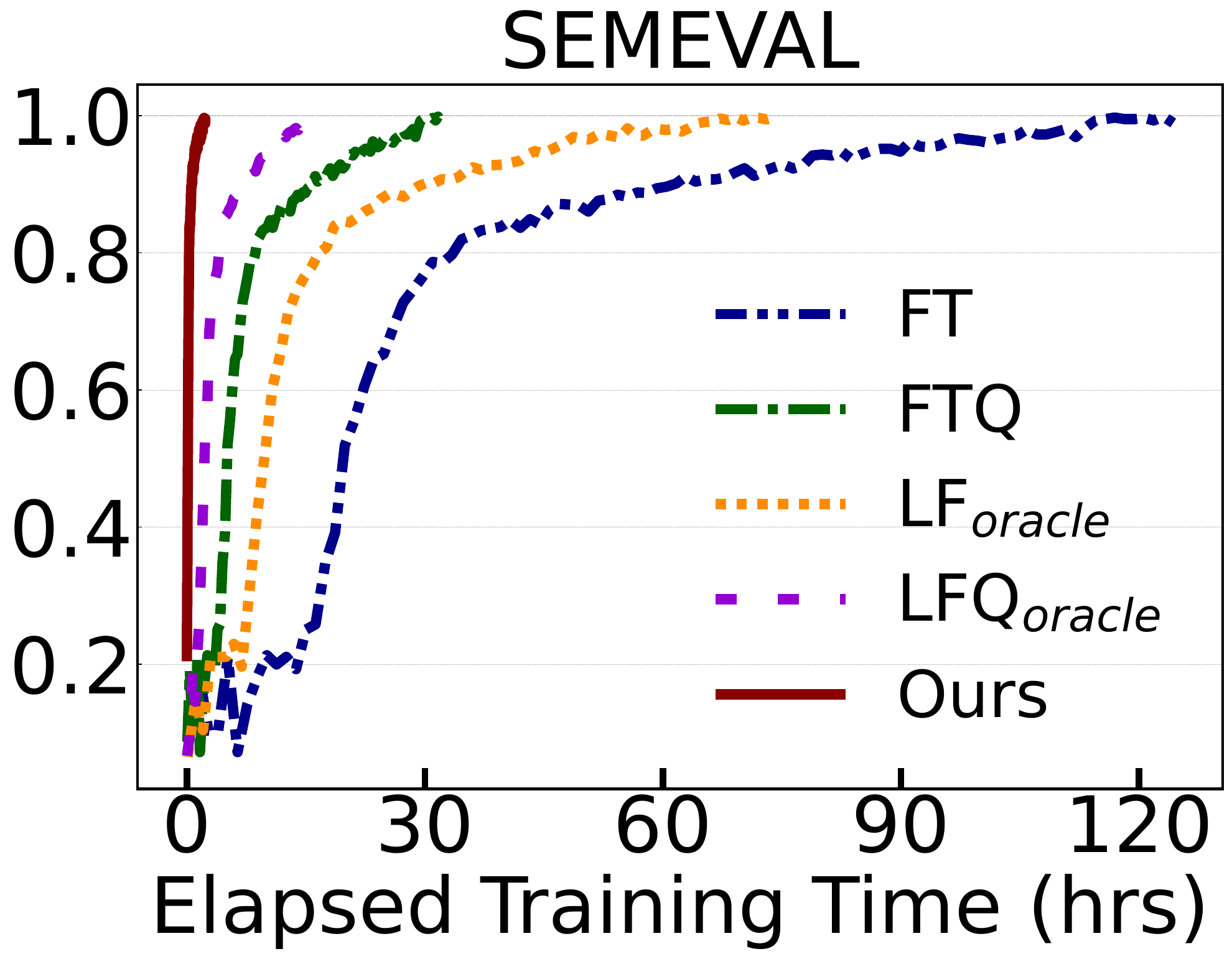}
        \end{minipage}
        ~
        \begin{minipage}[b]{0.24\textwidth}
            \includegraphics[width=0.95\textwidth]{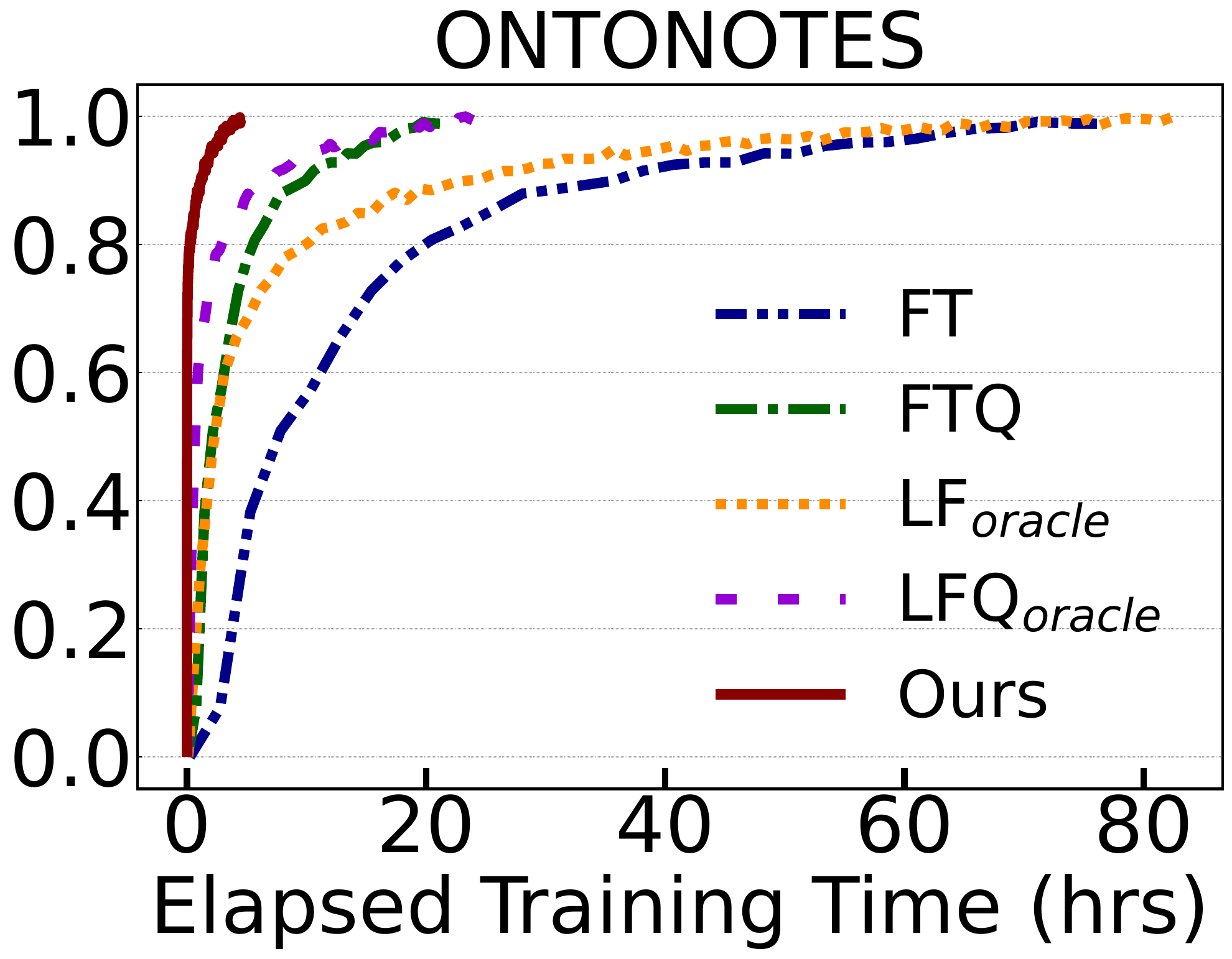}
        \end{minipage}
        \subcaption{BERT}\label{fig:eval-performance-bert}
    \end{minipage}
    
    \begin{minipage}[b]{1\textwidth}
        \begin{minipage}[b]{0.24\textwidth}
            \includegraphics[width=1.0\textwidth]{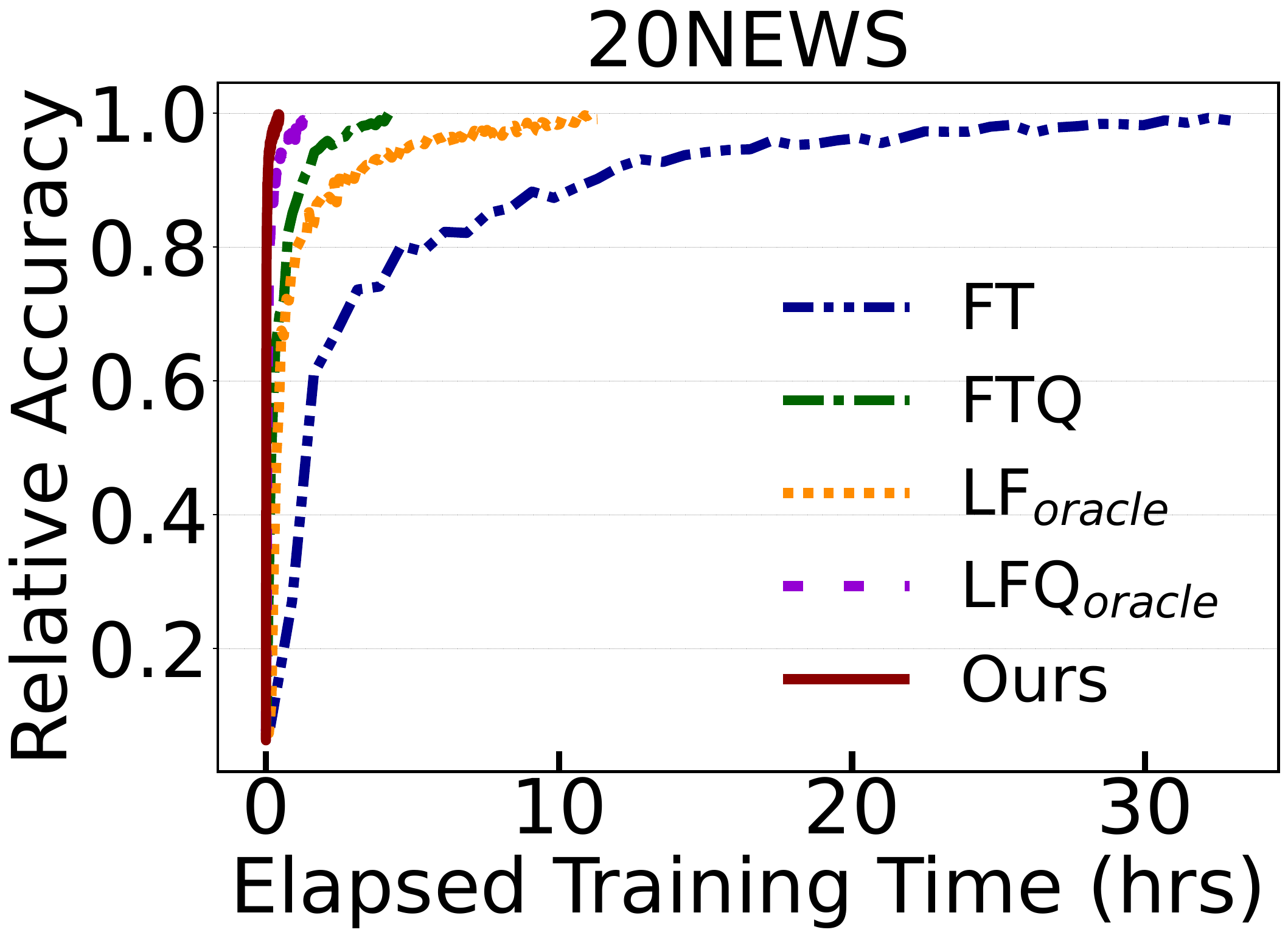}
        \end{minipage}
        ~
        \begin{minipage}[b]{0.24\textwidth}
            \includegraphics[width=0.95\textwidth]{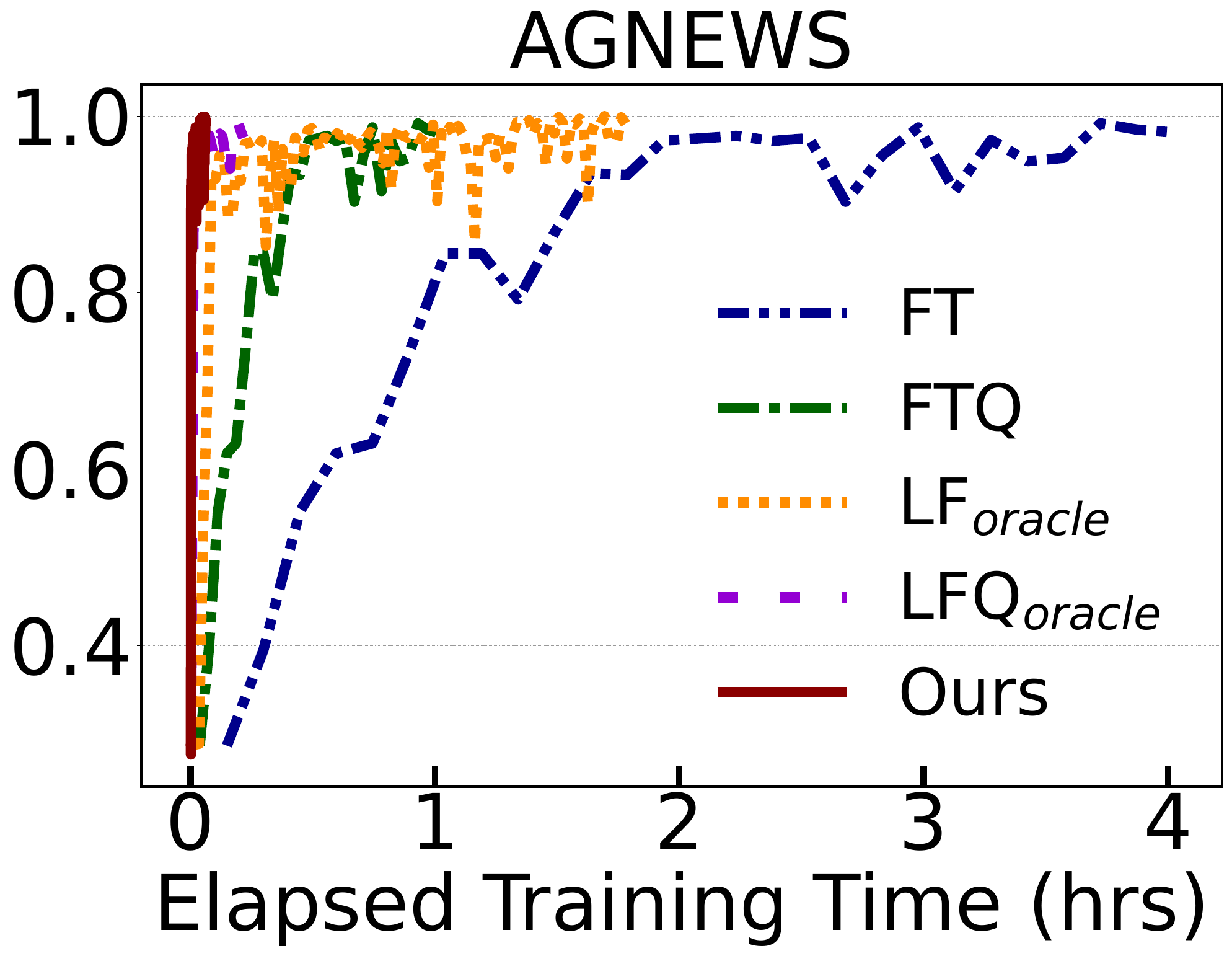}
        \end{minipage}
        ~
        \begin{minipage}[b]{0.24\textwidth}
            \includegraphics[width=0.95\textwidth]{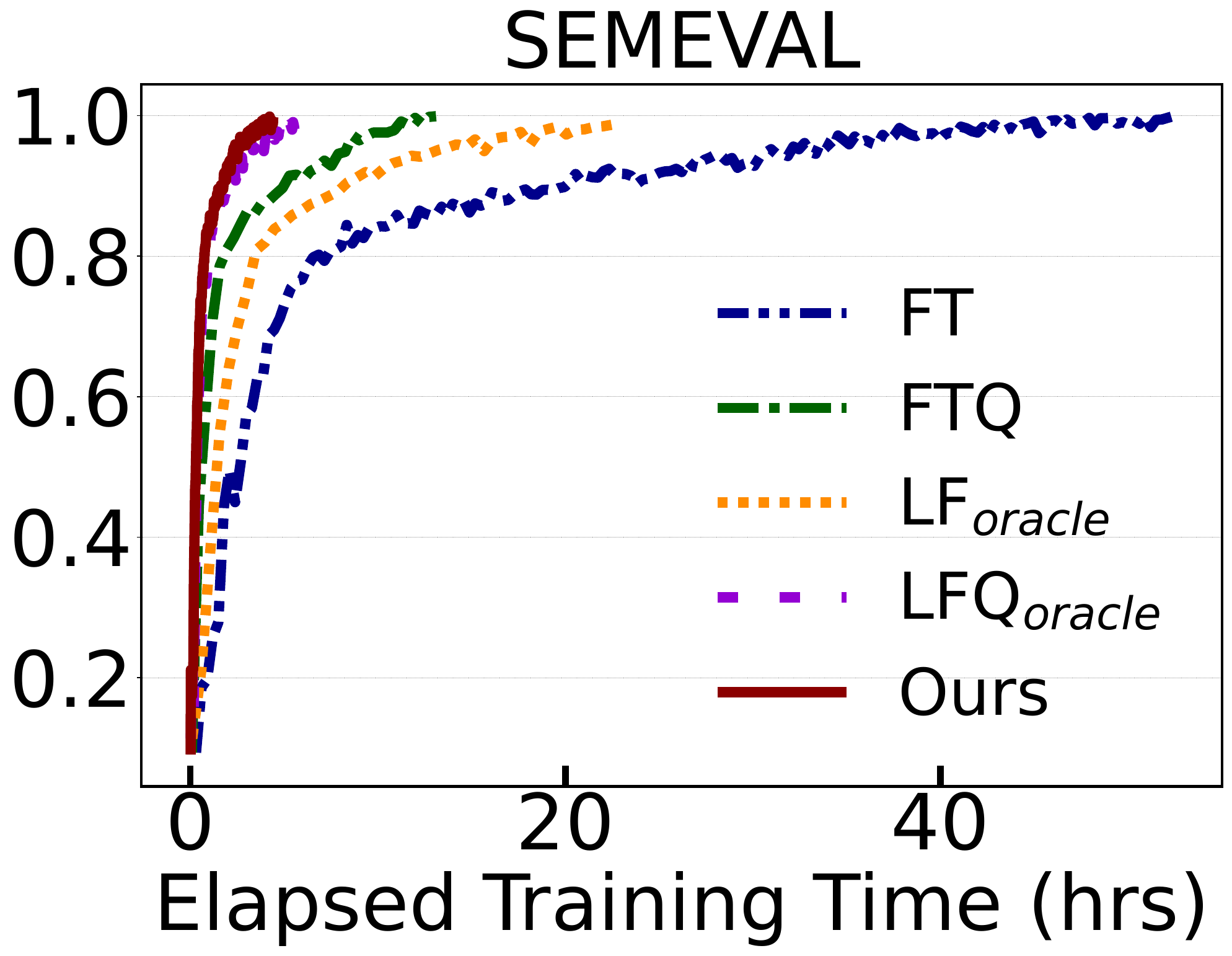}
        \end{minipage}
        ~
        \begin{minipage}[b]{0.24\textwidth}
            \includegraphics[width=0.95\textwidth]{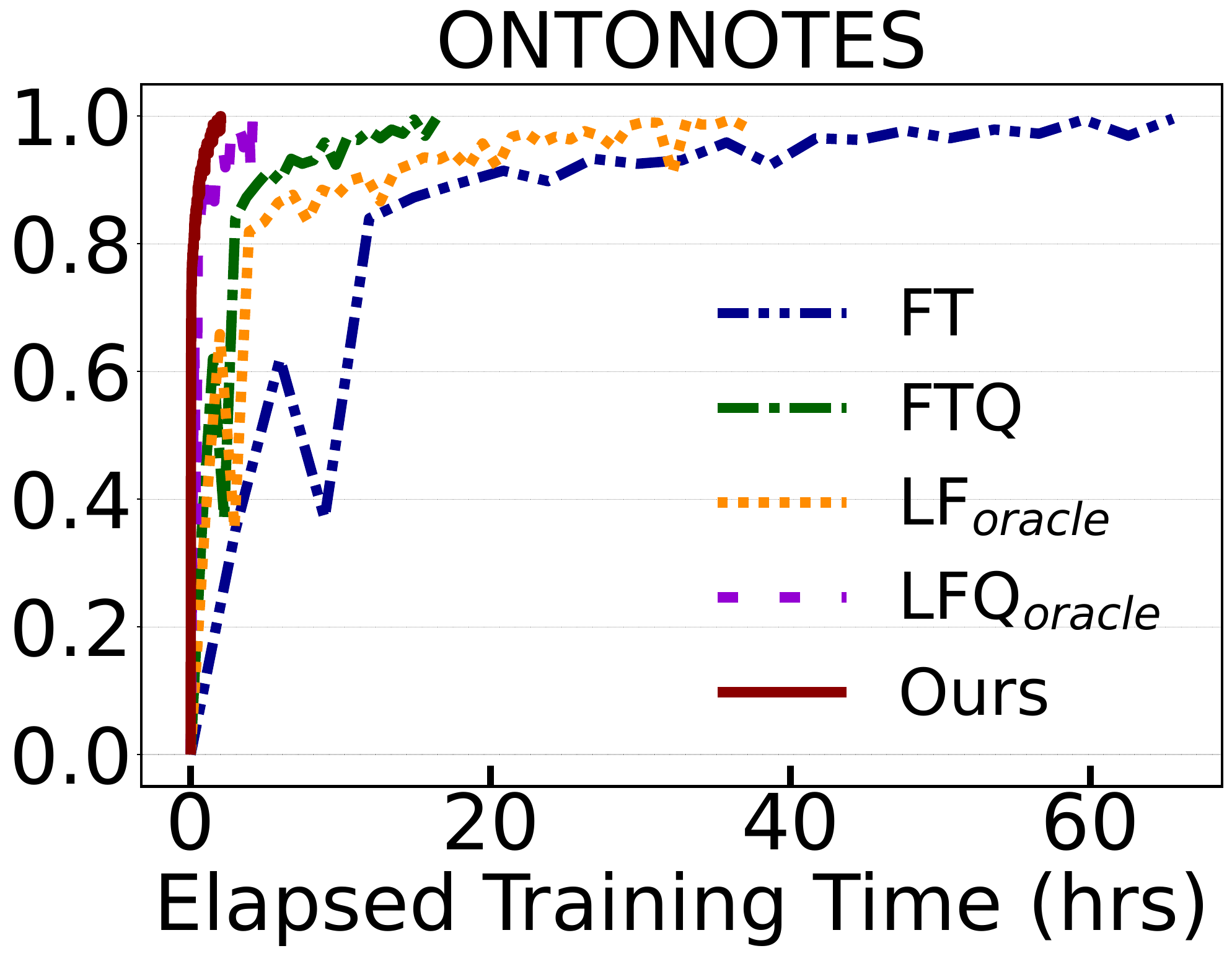}
        \end{minipage}
        \subcaption{DistilBERT}\label{fig:eval-performance-distilbert}
    \end{minipage}
    \caption{Time-to-accuracy throughout a training session. \sys{} speeds up model convergence significantly.}    
    
\end{figure*}

\textbf{Baselines} 
We compare \sys to the following alternatives.
(1) \texttt{Vanilla Fine-Tuning} (FT) always fine-tunes the whole model on each client.
This is the default fine-tuning methodology used in most NLP literature~\cite{devlin2018bert,sanh2019distilbert}.
(2) \texttt{Fine}-\texttt{Tuning}-\texttt{Quantized} (FTQ) quantizes the model parameters from FP32 to lower precision to reduce the network traffic between clients and aggregator.
Quantization is one of the most widely adopted approaches to reduce the communication cost and speedup FL process.
We use a state-of-the-art quantization algorithm~\cite{wu2018error} in our FedNLP tasks.
According to the algorithm, we observe the NLP models are quantized to INT4 or INT8 adaptively.
(3) \texttt{LayerFreeze-Oracle}~(\texttt{LF$_{oracle}$}) freezes a few transformer layers at bottom and only fine-tunes the ones above.
This is widely used to reduce the fine-tuning cost~\cite{guo2019spottune,lin2021fednlp}.
The number of freezed layers is selected per task to achieve the best time-to-accuracy at 99\% relative accuracy target.
(4) \texttt{LayerFreeze}-\texttt{Quantized}-\texttt{Oracle}~(\texttt{LFQ$_{oracle}$}) combines the above quantization and freezing techniques and selects the best setting for each task, i.e., the number of freezed layers and the quantized data precision.
To be noted, \texttt{LF$_{oracle}$} and \texttt{LFQ$_{oracle}$} are impractical in reality as they require prior knowledge to obtain an oracle system parameter.
For a fair comparison, all baselines use the same model aggregation algorithm (\textit{FedAvg}~\cite{mcmahan2017communication}) and client sampling (random), which are also the default setting in prior FL literature~\cite{lin2021fednlp}.

\textbf{Hyper-parameters}
Unless otherwise stated, \sys and all baselines use the same set of hyper-parameters as \texttt{FedNLP} \cite{lin2021fednlp} framework: mini-batch size as 4; local training iteration as 1; learning rate as 0.1; max sequence length as 256 for \texttt{20NEWS} and \texttt{ONTONOTES}, 64 for \texttt{AGNEWS} and \texttt{SEMEVAL}. 
For the FL configurations at the server side, we follow the prior FedNLP literature to select 15 participants by default for each training round, i.e., 5 clients in each trial group of \sys.

	\section{Evaluation}\label{sec:eval}
\subsection{End-to-end Performance}\label{sec:eval-e2e}


\textbf{\sys reduces model convergence delays significantly, making FedNLP practical.}
Table \ref{tab:runtime-performance} summarizes the convergence time and Figure \ref{fig:eval-performance-bert} illustrates the convergence process under the default setting.
To reach 99\% \relativeacc, \sys is 33.8$\times$, 155.5$\times$, 54.0$\times$ and 16.9$\times$ faster than \texttt{FT} on the four datasets, respectively.
With a lower target accuracy such as 90\%, the speedup brought by \sys is even more significant, i.e., 27.4$\times$--260.0$\times$.
It takes at most one hour for \sys to reach a usable accuracy.

More competitive baselines \texttt{LF$_{Oracle}$} and \texttt{FTQ} only bring limited improvement over \texttt{FT}, i.e., $2.4\times$--$3.9\times$ speedup.
Dozens of hours are still needed for a single downstream task.
\texttt{LFQ$_{Oracle}$} can benefit from both layer-freezing and quantization, therefore performing better than other baselines.
Though, \sys still beats \texttt{LFQ$_{Oracle}$} nontrivially, especially for reaching a relatively lower target accuracy.
For example, \sys is $12.8\times$ faster on \texttt{SEMEVAL} to reach 90\% \relativeacc.
This is because \sys employs an upgrading mechanism on adapter configuration which enables fast boosting of the training accuracy.
Note that both \texttt{LF$_{Oracle}$} and \texttt{LFQ$_{Oracle}$} are not practical methods as they use the ``optimal'' tuning depth which is not likely to be known beforehand.
If an improper depth is chosen, their performance drastically deteriorates.

We also extend our experiments to DistilBERT~\cite{sanh2019distilbert}, a distilled version of BERT, and illustrate the results in Figure \ref{fig:eval-performance-distilbert}.
It shows that \sys significantly outperforms the baselines on DistilBERT as well.
For instance, \sys achieves $14.89\times$--$73.42\times$ speedup over \texttt{FT} to obtain the 99\% \relativeacc.
Interestingly, comparing DistilBert to BERT under \texttt{FT}, we find the former achieves up to $7.7\times$ speedup on \texttt{AGNEWS} on Jetson TX2 since it is more lightweight in consideration of both computation and communication.
However, this advantage does not hold for all circumstances.
Comparing Figure \ref{fig:eval-performance-distilbert} (d) with Figure \ref{fig:eval-performance-bert} (d), we find that BERT has comparable performance with DistilBERT on \texttt{ONTONOTES}.
It is likely attributed that, when the downstream task gets harder, the loss of model's representation capacity during distilling incurs negative impacts.
Such a difference will compensate the large model size and computation complexity of the vanilla BERT.
Nevertheless, \sys performs consistently on both models because it judiciously tunes the pluggable adapters with different depths and widths.

\begin{figure}[t]
	\centering
	\vspace*{-4pt}\hspace*{3pt}\begin{minipage}[b]{0.24\textwidth}
        \hspace*{-3pt}\includegraphics[width=1.9\textwidth]{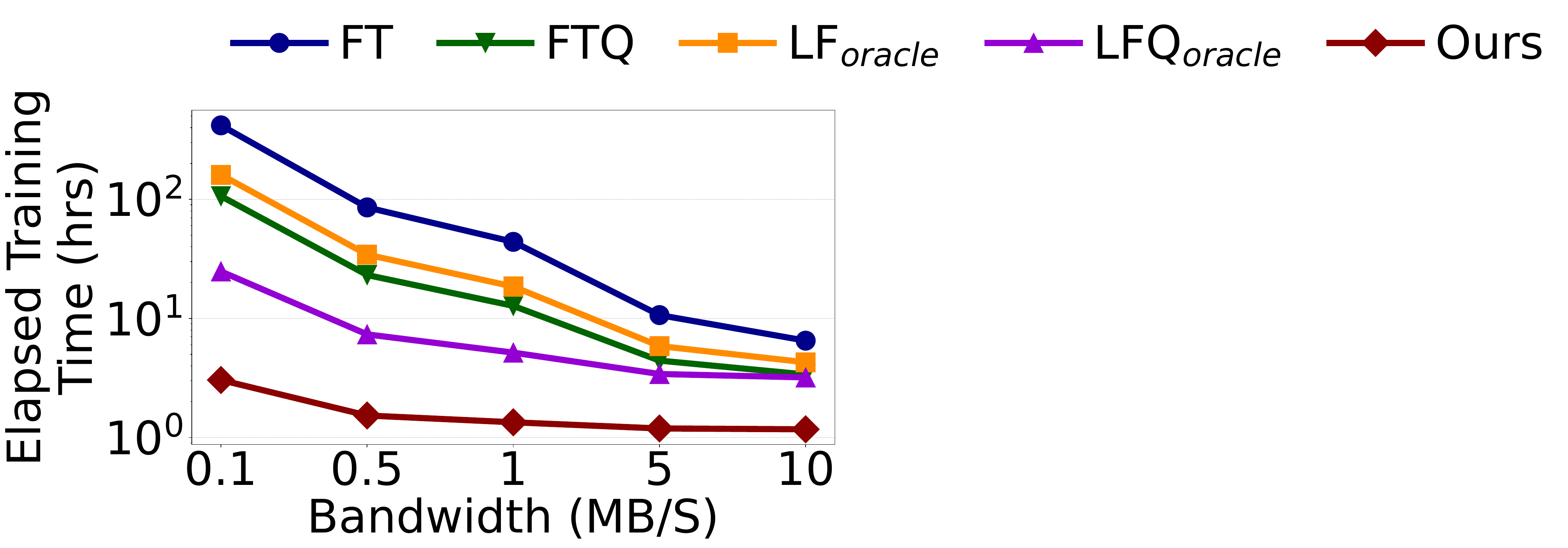}\vspace*{-4pt}
        \subcaption{\texttt{20NEWS}}
    \end{minipage}
	~
    \hspace*{-14pt}\begin{minipage}[b]{0.24\textwidth}
        \hspace*{10pt}\includegraphics[width=0.88\textwidth]{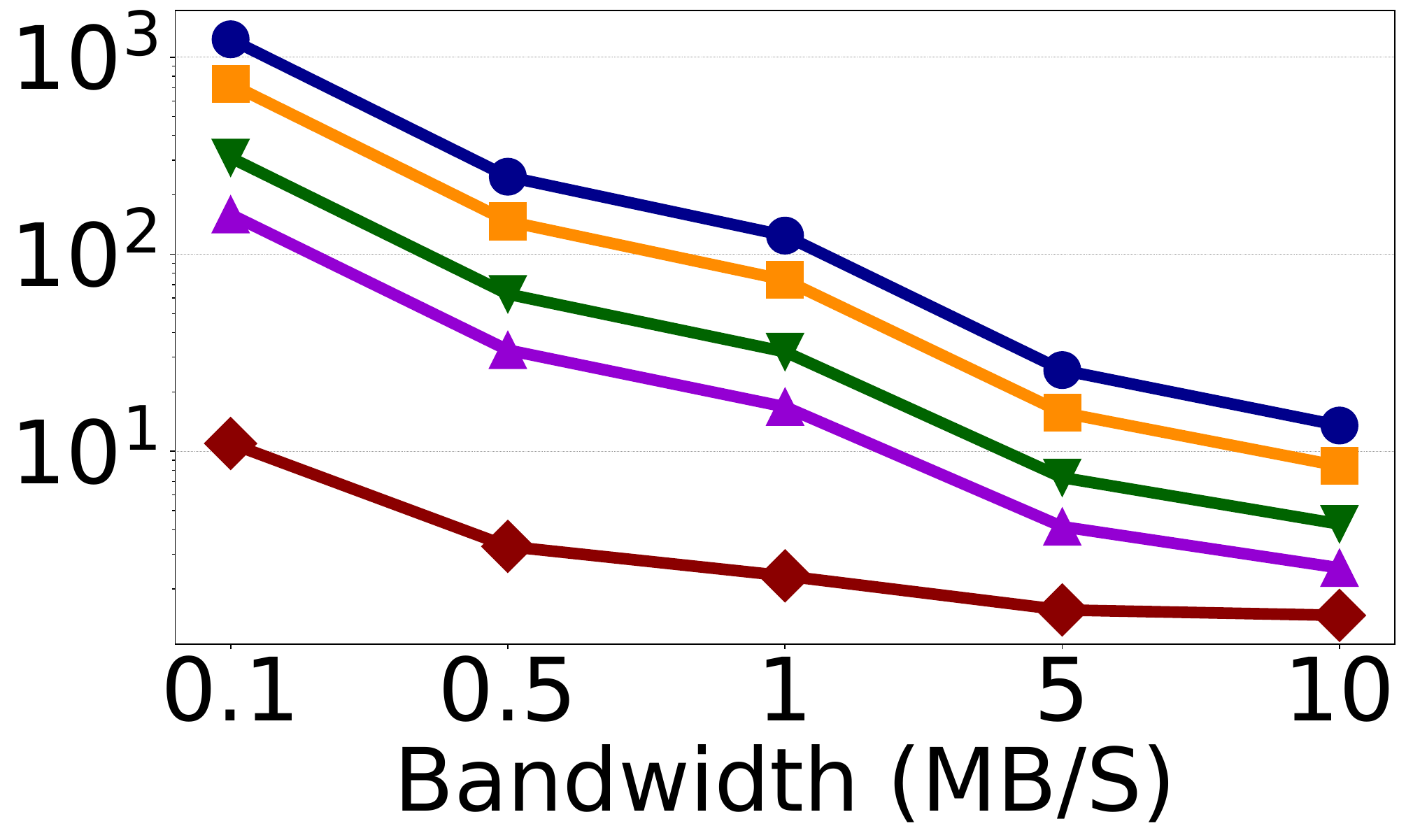}\vspace*{-2.5pt}
        \subcaption{\texttt{SEMEVAL}}
    \end{minipage}
	\caption{
	\sys{} outperforms baselines under all network bandwidths with 99\% target accuracy.
	}
	\label{fig:eval-performance-bandwidth}
\end{figure} 

\textbf{\sys{} outperforms baselines in various network environments.}
Figure \ref*{fig:eval-performance-bandwidth} reports the performance of \sys and baselines under various network environment from 0.1MB/s to 10MB/s, which cover the typical network capacity for nowaday WiFi and cellular bandwidth.
Our key observation is that \sys consistently outperforms other baselines with different network conditions, and the improvement is more significant with lower network bandwidth.
For instance, with $bandwidth=10MB/s$, \sys reaches the 99\% \relativeacc $5.6\times$ and $9.2\times$ faster than \texttt{FT} on \texttt{20NEWS} and \texttt{SEMEVAL}, seperately.
When the bandwidth goes down to 0.1MB/s, the improvement is as high as 137.7$\times$ and 112.5$\times$, respectively.
The rationale behind this is that \sys brings the most network transmission reduction by inserting tiny adapter modules into the model.
Such a micro transmission package makes the communication process fast even with very low network bandwidth.
In reality, network fluctuation is common~\cite{xu2020client}.
\sys enables a stable fine-tuning process and paves the way for the fundamental solution to stragglers or other possible communication problems~\cite{reisizadeh2020straggler,wang2021device} that will drag the NLP fine-tuning slow.

\begin{figure}[t]
	\centering
	\vspace*{-5pt}\begin{minipage}[b]{0.48\textwidth}
        \includegraphics[width=1.0\textwidth]{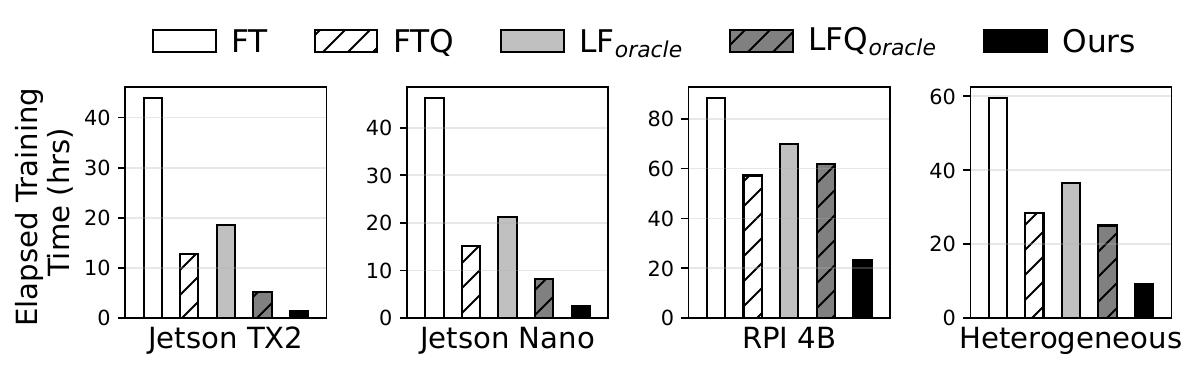}
        \subcaption{\texttt{20NEWS}}
    \end{minipage}

    \begin{minipage}[b]{0.48\textwidth}
        \includegraphics[width=1.0\textwidth]{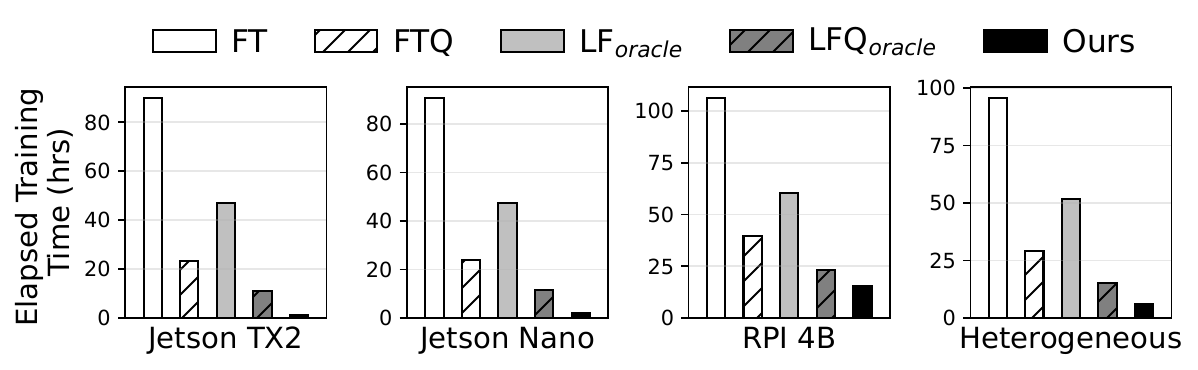}
        \subcaption{\texttt{SEMEVAL}}
    \end{minipage}

	\caption{Convergence delays with a variety of client hardware.
	Training targets 99\% \relativeacc.
	``Heterogeneous'' means the device capacity is uniformly distributed between three boards.}
	
	\label{fig:eval-device-heterogeneity-performance}
\end{figure}
\textbf{\sys{} outperforms baselines on various client hardware.}
\sys also consistently outperforms other baselines with different device capacity as shown in Figure~\ref{fig:eval-device-heterogeneity-performance}.
On GPU-powered high-end embedded devices like Jetson TX2 and Jetson Nano, \sys reaches the 99\% \relativeacc up tp 32.8$\times$ and 79.2$\times$ faster than the \texttt{VanilaFT} on \texttt{20NEWS} and \texttt{SEMEVAL}, respectively.
On a much wimpy device RPI 4B, the speedup degrades to 3.8$\times$ and 7.0$\times$, respectively.
This is because \sys reduces two orders of magnitude in communication cost, but only one order of magnitude in computation cost. 
Nevertheless, \sys can still bring remarkable improvement on the wimpy devices.
For example, on dataset \texttt{20NEWS}, \sys reaches target accuracy $5.03\times$ and $3.25\times$ faster than the four baselines, respectively.
Moreover, \sys also shows improvement under  heterogeneous settings, where the device capacity is assumed to be uniformly distributed between Jetson TX2, Jetson Nano and RPI 4B.
The profit is not as significant as on high-end devices because the per-round training is bottlenecked by the slower devices~\cite{reisizadeh2020straggler}.
Extensive research efforts have been invested to mitigate straggler issue~\cite{li2021hermes,lipyramidfl, nishio2019client, xu2020client, wang2021device, lai2020oort, zhao2021quality} and \sys is orthogonal to those techniques.

\subsection{Significance of Key Designs}
\label{sec:eval-ablation}


The benefits of \sys come from: the adapters ($\S$\ref{sec:design:adapter}), the activation cache ($\S$\ref{sec:design:cache}), the automatic adapter configuration and trial-and-error clients ($\S$\ref{sec:design:scheduler}).
We now quantify their benefits. 

\begin{figure}[t]
	\centering
	
	\includegraphics[width=0.49\textwidth]{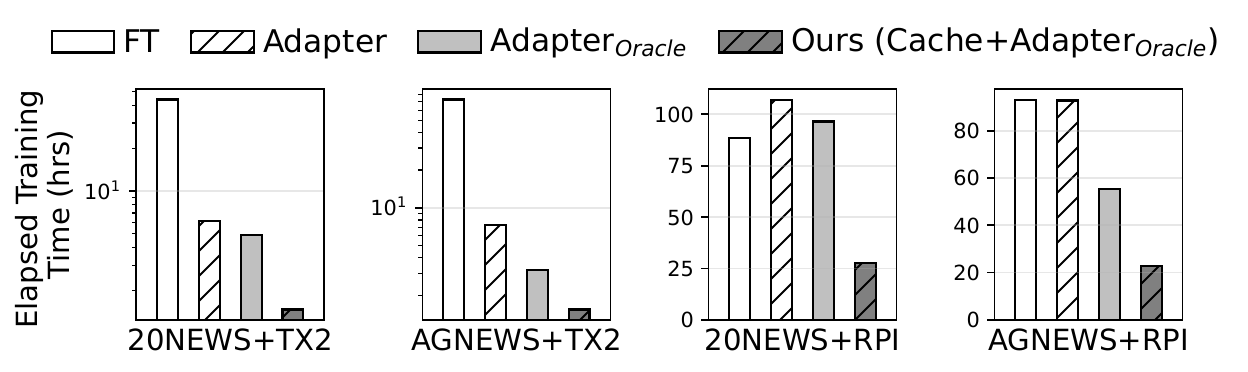}
	\caption{Model convergence delays with and without \sys{}'s key designs, showing their significance.} 
    \label{fig:eval-ablation-adapter-cache}
\end{figure}

\begin{figure}[t]
	\centering
	\begin{minipage}[b]{0.22\textwidth}
		\hspace{-6pt}\includegraphics[width=1.05\textwidth]{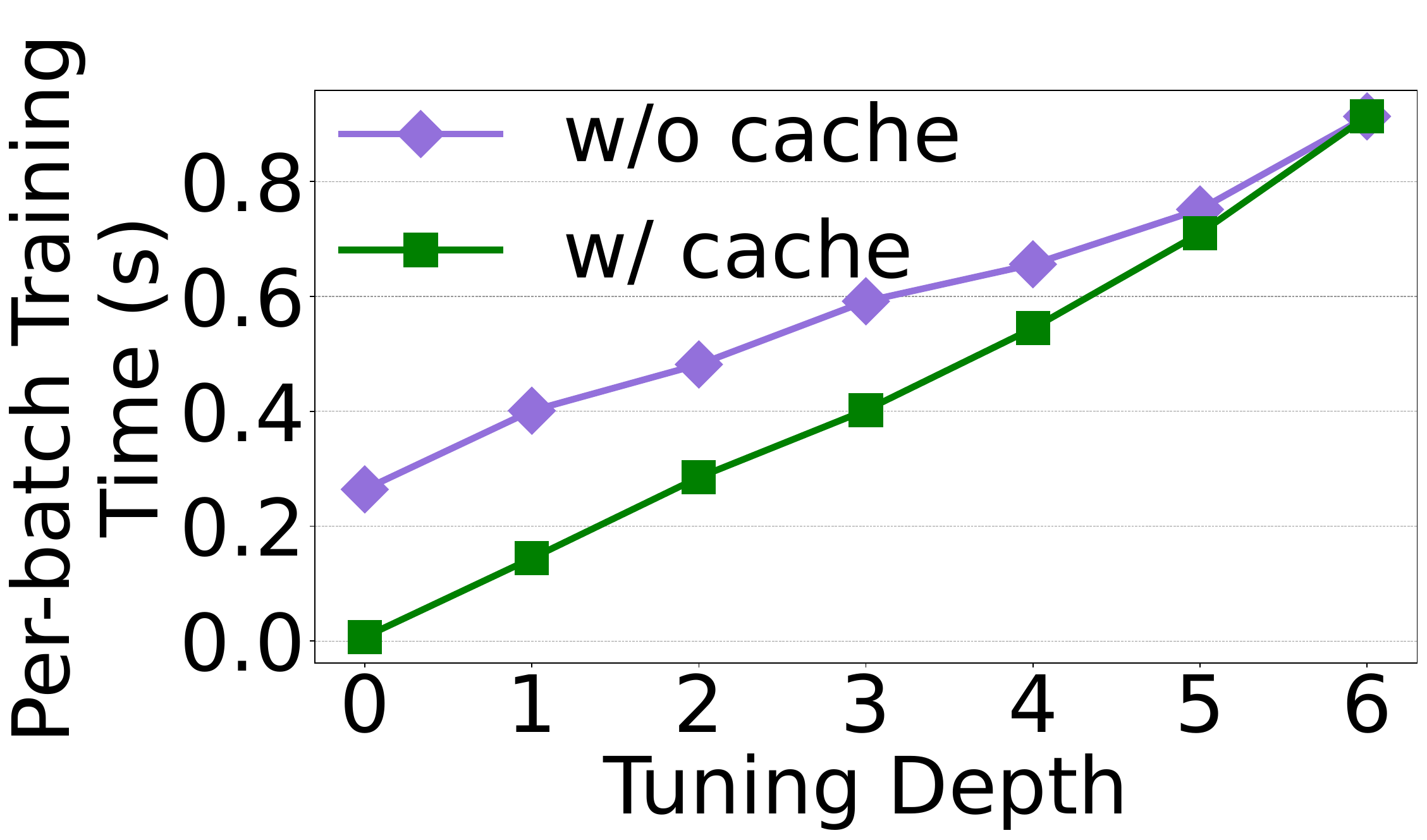}
		\caption{Per-batch training time with/without activation caching. Model: DistilBERT. Device: Jetson TX2.} 
		\label{fig:design-cache-perf}
	\end{minipage}
	\hspace{5pt}
	\begin{minipage}[b]{0.22\textwidth}
        \includegraphics[width=1\textwidth]{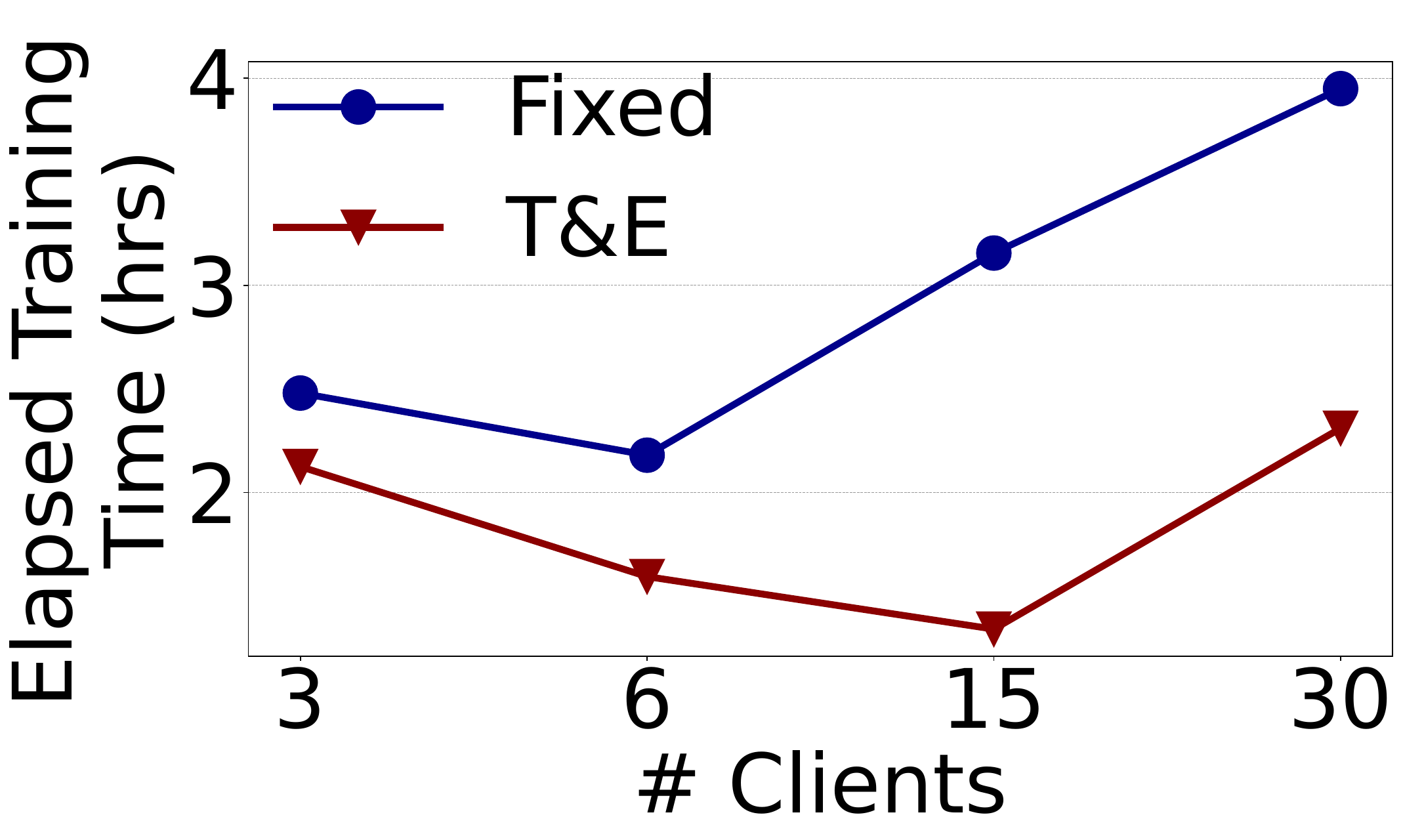}
		\caption{
			Model converge time with different client numbers.
		Dataset: \texttt{20NEWS}. Model: BERT. Device: Jetson TX2.}
		\label{fig:eval-ablation-clients}
    \end{minipage}
\end{figure}

\textbf{Adapter and caching.}
Figure \ref{fig:eval-ablation-adapter-cache} shows benefit of adapters and  caching. 
Naive use of adapters, e.g. inserting at all layers, may brings notable benefit, 
e.g., on Jetson TX2 and benchmark \texttt{AGNEWS}, \texttt{Vanilla-Adapter} is 10.1$\times$ faster than \texttt{FT}. 
However, the delays are still too high.
On RPI 4B, naive use of adapters slows down the model convergence as compared to fine-tuning the whole model. 
By using one static, oracle adapter configuration (\texttt{Adapter$_{oracle}$}), the time-to-accuracy is reduced by up tp 23.0$\times$ compared to \texttt{FT}.
Employing the activation cache technique (\texttt{Adapter$_{oracle}$+Cache}) further brings 2.1$\times$--3.3$\times$ speedup.
In another micro experiment, Figure~\ref{fig:design-cache-perf} shows that, with activation caching, the training time decreases almost linearly with fewer adapter layers to be updated ($D-d_{split}$), and the improvement from caching mechanism increases significantly as well.

\begin{figure}[t]
	\centering
	\begin{minipage}[b]{0.23\textwidth}
		\centering
		\includegraphics[width=1\textwidth]{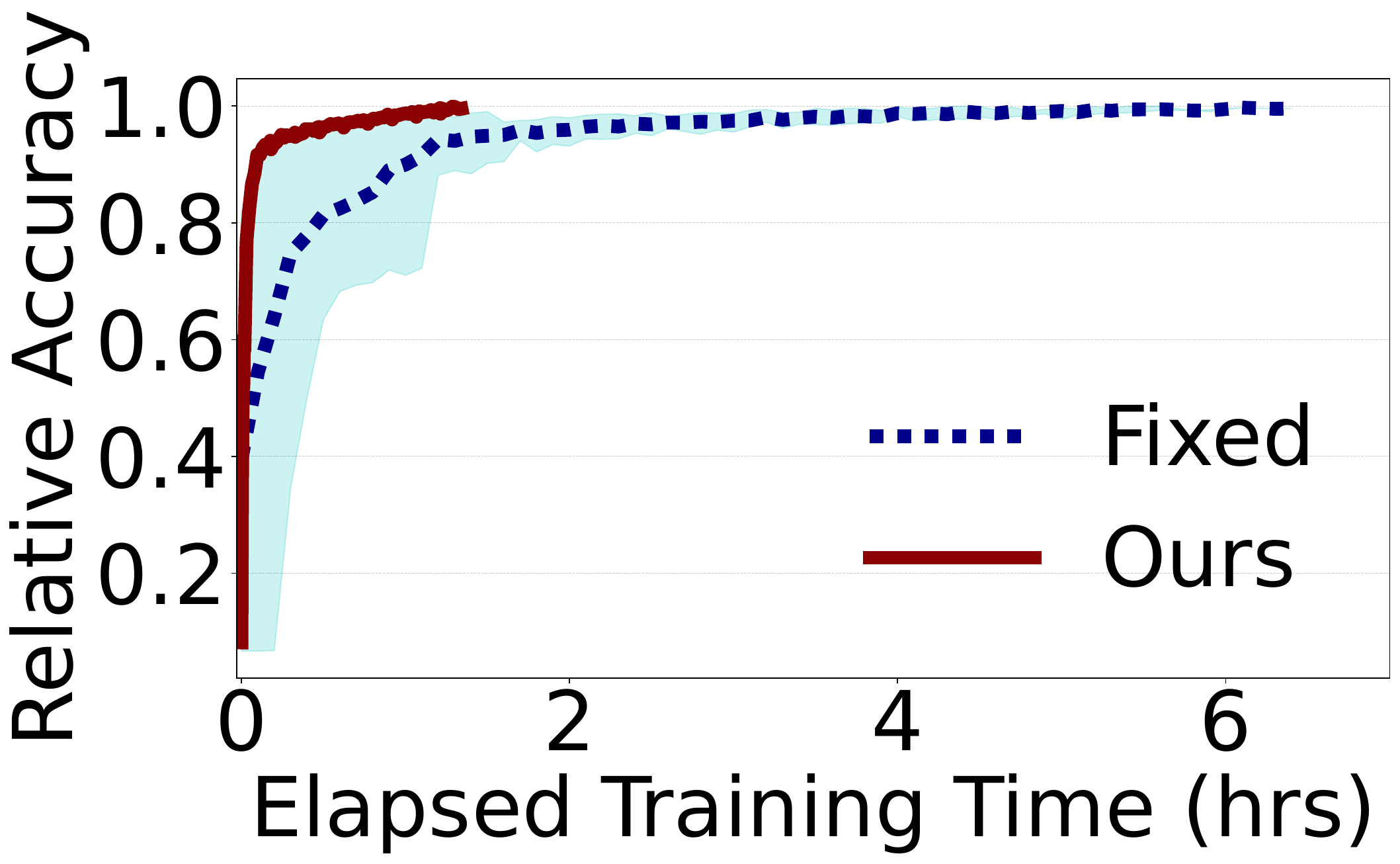}
		\subcaption{\texttt{20NEWS}}
		\label{fig:eval-ablation-adaptive}
	\end{minipage}
	\begin{minipage}[b]{0.23\textwidth}
		\centering
		\includegraphics[width=0.96\textwidth]{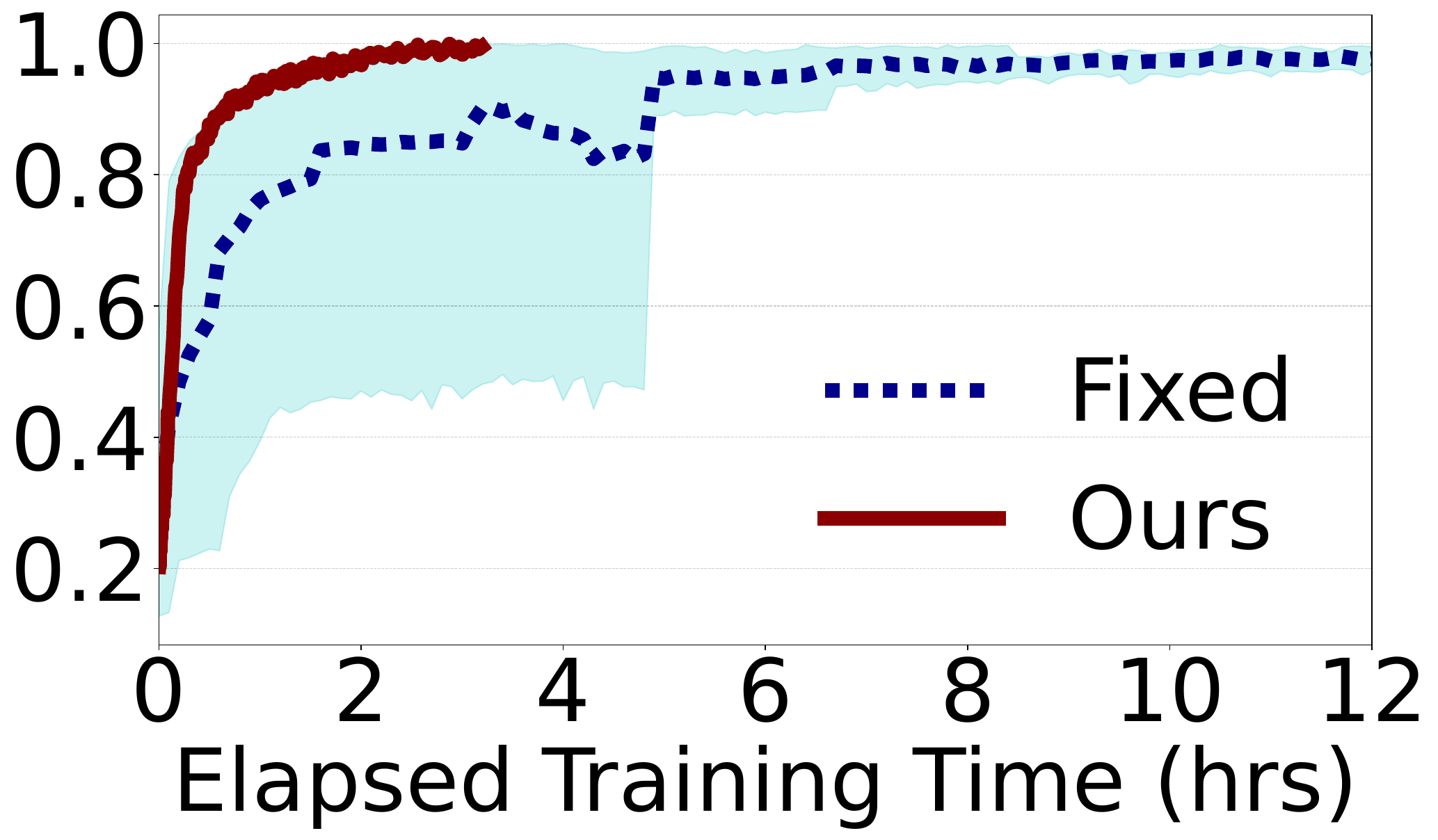}
		\subcaption{\texttt{SEMEVAL}}
	\end{minipage}
	\caption{Time-to-accuracy throughout a training session. \sys{}'s accuracy (red lines) always outperforms those of fixed adapter configuration (208 in total, aggregated as blue shades, for which blue dotted lines show averages).}
	\label{fig:eval-ablation-adaptive}
\end{figure}

\textbf{Automatic configuration}
To demonstrate the importance of \sys's upgrading mechanism on the adapter's tuning configuration, we exhaustively sweep through all adapter configurations (depth 0--12, width 8,16,..,128, 208 configurations in total) of BERT on \texttt{20NEWS}, and aggregate their convergence curves as shaded areas shown in Figure~\ref{fig:eval-ablation-adaptive}.
The blue line (dotted) is the average time-to-accuracy of all configurations while the red line (solid) is the curve of \sys{}.
Note that sweeping all configurations is very expensive: it takes thousands of GPU hours to run the benchmark in a subfigure.
The results show that \sys almost outperforms every configuration throughout a training session. 
This is owing to \sys{} switching among different configurations that best suits the current training session.
Typically, we observe \sys uses 8--14 configurations per training session.

\textbf{Investment of extra clients}
\sys uses more clients to identify whether it shall upgrade to a more complex adapter configuration through trial and error.
We evaluate the performance of \sys with different client numbers as compared to \texttt{VanillaFT}.
As show in Figure~\ref{fig:eval-ablation-clients}, \texttt{VanillaFT} achieves the best performance with 6 participant clients per round.
On the other hand, using the extra clients for trial-and-error is much more beneficial, i.e., better scalability to the available clients.


\subsection{Client Resource Cost} 
\label{sec:eval-cost}


\begin{figure}[t]
	\centering
	\includegraphics[width=0.48\textwidth]{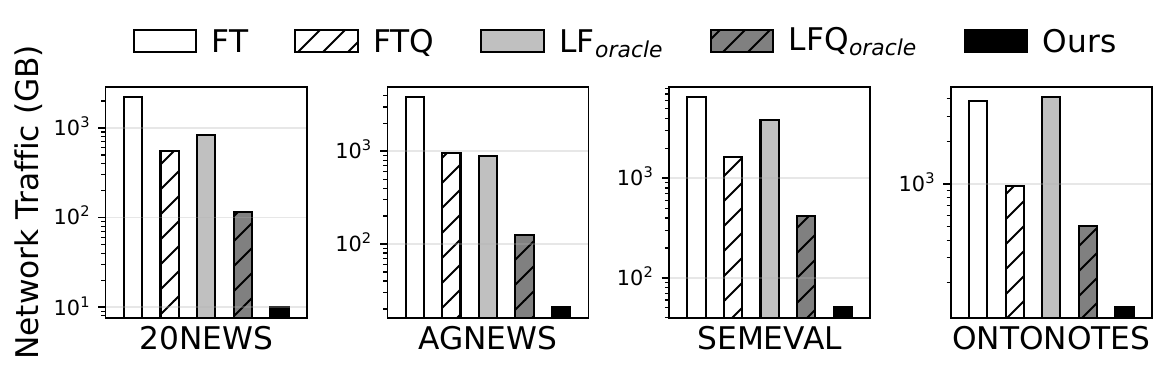}
	\caption{Total network traffic of all client devices. Training targets 99\% \relativeacc{}.}
    \label{fig:eval-network}
\end{figure}

\textbf{Network traffic.}
Figure \ref{fig:eval-network} reports the total network traffic incurred during fine-tuning to reach 99\% \relativeacc. 
It shows that \sys saves 126.7$\times$ on average and up to $220.7\times$ (reducing from 2194.3 GB to 9.9 GB) network traffic compared to the \texttt{FT} on dataset \texttt{20NEWS}.  
Note that reducing the network traffic not only speeds up the convergence, 
but also mitigates the overhead on clients and the monetary cost to FL developers, which is billed by the amount of data transmitted on public cloud platforms, e.g., \$0.01/GB on AWS~\cite{awsbill2022}.

\begin{figure}[t]
	\centering
	\includegraphics[width=0.48\textwidth]{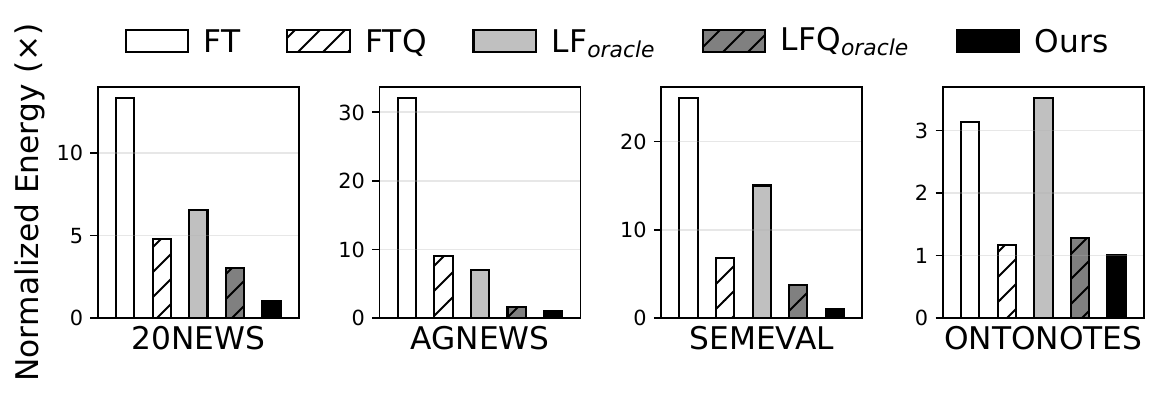}
	\caption{Per-client average energy consumption, normalized to that of \sys. 
	Training targets 99\% \relativeacc.
}
    \label{fig:eval-energy}
\end{figure}

\textbf{Energy consumption.}
Figure~\ref{fig:eval-energy} illustrates the average energy consumed during FedNLP tasks on each device.
It shows that \sys saves the energy consumption remarkably, e.g., $1.3\times$--$3.7\times$ reduction compared to \texttt{LFQ$_{Oracle}$} and $3.1\times$--$32.1\times$ reduction compared to \texttt{FT}, respectively.
Such improvement comes from both the reduced network transmission time and the on-device training computations.

\begin{figure}[t]
	\centering
    \hspace*{15pt}
	\begin{minipage}[b]{0.24\textwidth}
        \hspace*{-4pt}\includegraphics[width=1.65\textwidth]{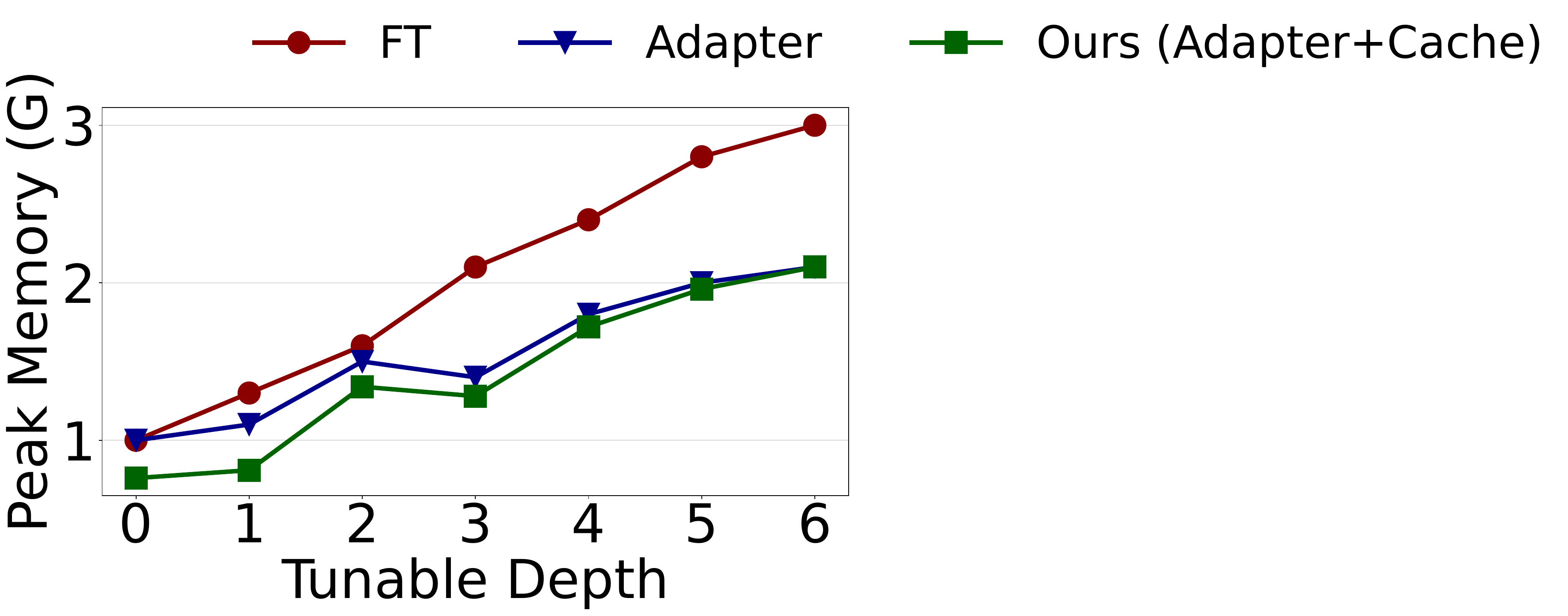}
        \subcaption{DistilBERT}
    \end{minipage}
	~
    \hspace*{-30pt}
	\begin{minipage}[b]{0.24\textwidth}
        \hspace*{10pt}\includegraphics[width=0.83\textwidth]{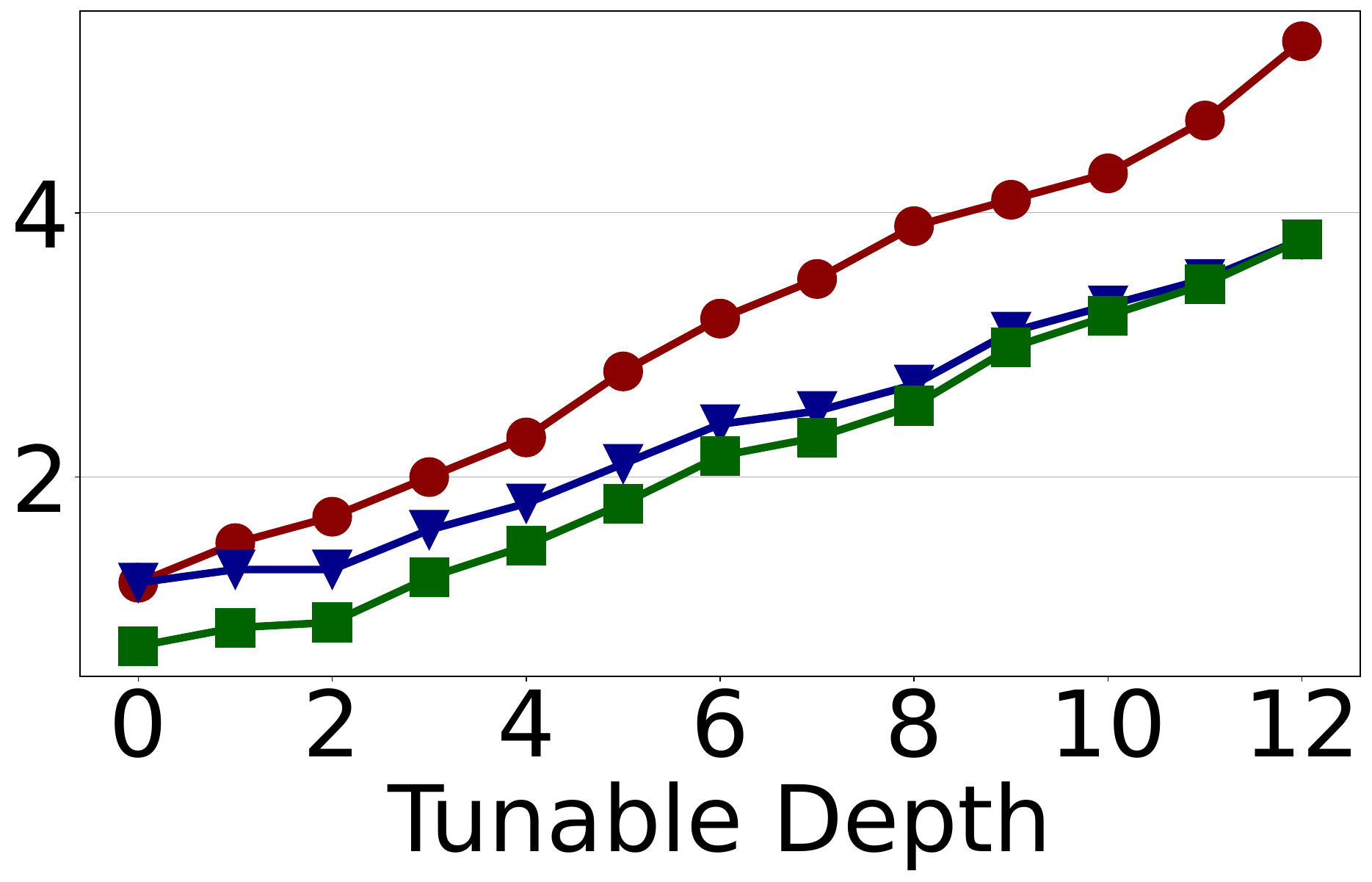}\vspace*{0.6pt}
        \subcaption{BERT}
    \end{minipage}
	\caption{Peak memory usage of a client device. 
	}
	\label{fig:eval-memory}
\end{figure}

\textbf{Memory footprint}
Figure \ref{fig:eval-memory} reports the peak memory footprint when fine-tuning on BERT and DistilBERT with different tuning depths.
It shows that \sys nontrivially reduces the memory usage either with shallow or deep tuning depth.
The reasons are twofold.
First, \sys only updates the parameters of a few adapters, so the gradients of other parameters are not calculated and the associated activations do not need to be stored.
Second, our activation caching technique avoids storing the unneeded parameters.





	\section{Related Work}\label{sec:related}

\paragraph{Fine-tuning (transfer learning)}
Inductive transfer learning has greatly advanced NLP research.
Howard et al. propose ULMFiT~\cite{howard2018universal}, a universal transfer learning method matching the performance of training from scratch.
BERT~\cite{devlin2018bert} was then introduced and becomes a standard pre-trained model in many NLP downstream tasks for its superior performance and generality.
Numerous variants~\cite{vaswani2017attention,devlin2018bert,sanh2019distilbert, hou2020dynabert, liu2020fastbert, sun2020mobilebert, zafrir2019q8bert, bai2020binarybert} of BERT have since been designed.
For instance, Sun et al. explore the space of strategies for fine-tuning BERT for text classification~\cite{sun2019fine}. 
This work is motivated by those work and specifically targets FedNLP scenario.

\paragraph{FedNLP} is a key step towards the adoption of NLP models in practice.
However, there is very few literature investigating its implications at system aspect.
\cite{lin2021fednlp} is the first research benchmark for FedNLP tasks and integrates representative language datasets.
\sys is built atop it and treats it as a baseline.
SEFL~\cite{deng2022secure} is a FedNLP framework that achieves data privacy without any trusted entities.
\cite{basu2021benchmarking} studies how FedNLP can orchestrate with differential privacy.
None of above work addresses the high training cost of FedNLP.

\paragraph{Adapters}
Adapter is extensively studied to achieve parameter efficiency in continuous learning tasks. 
It was first introduced for vision tasks~\cite{rebuffi2017learning}. 
The rationale is to encode task-specific representations in intermediate layers while preserving the knowledge learned from the pre-training dataset~\cite{pfeiffer2021adapterfusion}.
Various adapter variants have been proposed to tradeoff trainable parameter numbers and training accuracy in NLP tasks~\cite{pfeiffer2020mad, li2021prefix, mahabadi2021compacter, he2021towards, sung2021training}.
Despite the popularity, the implications of adapter in FedNLP tasks have not been well examined.
For the first time, we treat adapter as a building block to address the training performance issue in FedNLP.

\paragraph{Optimizations for FL}
Due to the decentralized nature, communication has been recognized as a major bottleneck in FL tasks~\cite{bonawitz2019towards,yang2021characterizing}.
Various optimizations~\cite{wang2020intermittent, li2021sample, wu2018error, bernstein2018signsgd, wangni2018gradient} have been proposed.
Among them, model compression/quantization~\cite{wu2018error, bernstein2018signsgd} is the mostly adopted and is directly compared in this work.
Apart from network transmission, data and device heterogeneity~\cite{reddi2020adaptive} are also unique challenges introduced in FL.
To mitigate the heterogeneity of client devices (therefore stragglers), Abdelmoniem et al.~\cite{abdelmoniem2021towards} ask each client device to quantize their local model adaptively.
Hermes~\cite{li2021hermes} guides different mobile clients to find a small subnetwork through structured pruning for local training.
Most existing vision-based FL optimizations are CNN-specific~\cite{li2021hermes,he2020group} and thus cannot apply to FedNLP.
Some of them are compatible with \sys, e.g., intelligent client selection and data sampling~\cite{li2021hermes,lipyramidfl, nishio2019client, xu2020client, wang2021device, lai2020oort, zhao2021quality, li2021sample}.
\sys instead takes the first fundamental step towards practical FedNLP, and is compatible with above techniques.

	\section{Conclusions}\label{sec:conclusions}

\sys is a federated learning framework for fast NLP model fine-tuning.
\sys borrows the wisdom from prior work and uses adapter as the only trainable module in NLP model to reduce the training cost.
To identify the optimal adapter configuration on the fly, \sys integrates a progressive training paradigm and trail-and-error profiling technique.
\sys shows superior training speedup over existing approaches through our extensive experiments.

\section*{Acknowledgments}

This research was supported by National Key Research and Development Program of China \#2020YFB1805500, the Fundamental Research Funds for the Central Universities, and NSFC \#62032003, \#61922017, \#61921003. 
Mengwei Xu was partly supported by NSFC \#62102045, Beijing Nova Program \#Z211100002121118, and Young Elite Scientists Sponsorship Program by CAST \#2021QNRC001. 
The authors thank the anonymous reviewers and the shepherd for their insightful feedbacks.

	\bibliographystyle{plain}
	\bibliography{bib/ref-mwx,bib/nlp,bib/ref-cdq}

\end{document}